\documentclass[letterpaper, 10pt, conference]{ieeeconf}      

\IEEEoverridecommandlockouts                              
\renewcommand{\baselinestretch}{0.94}

\usepackage{nicefrac}
\usepackage{url}
\usepackage{subfig}
\usepackage[dvipsnames]{xcolor}
\usepackage{multirow}
\usepackage{bbold}
\usepackage{balance}

\usepackage{breakurl}

\usepackage{footnote}
\makesavenoteenv{tabular}
\makesavenoteenv{table}

\def\HiLi{\leavevmode\rlap{\hbox to \hsize{\color{yellow!20}\leaders\hrule height .8\baselineskip depth .5ex\hfill}}}
\def\HiLiGreen{\leavevmode\rlap{\hbox to \hsize{\color{green!20}\leaders\hrule height .8\baselineskip depth .5ex\hfill}}}
\def\HiLiBlue{\leavevmode\rlap{\hbox to \hsize{\color{blue!20}\leaders\hrule height .8\baselineskip depth .5ex\hfill}}}
\def\HiLiGray{\leavevmode\rlap{\hbox to \hsize{\color{grey!50}\leaders\hrule height .8\baselineskip depth .5ex\hfill}}}

\usepackage{epsfig}
\usepackage{amssymb}
\usepackage{amsmath}
\usepackage{amsfonts}

\usepackage{stackengine}
\usepackage{algorithmic,comment}
\usepackage[ruled,vlined,commentsnumbered,titlenotnumbered,linesnumbered]{algorithm2e}
\usepackage{float}

\usepackage{colortbl}
\usepackage{xcolor}
\definecolor{LightCyan}{rgb}{0.88,1,1}

\usepackage{color}
\newcommand{\SC}[1]{{\color{black}#1}}
\newcommand{\revision}[1]{{\color{black}#1}}

\newcommand{\tu}[1]{{\underline{\textit{#1}}}}

\newcommand{\name}{\textsc{TaskNet}\xspace}
\newcommand{\MPNET}{\textsc{MPNet}\xspace}
\newcommand{\lossname}{uncompressed input task loss}

\newcommand{\reals}{\mathbb{R}}

\newcommand{\Nepochs}{T}

\newcommand{\argmin}{\mathop{\rm argmin}}

\newtheorem{thm}{Theorem}

\newtheorem{problem}{Problem}

\newcommand{\yhat}{\hat{y}}

\newcommand{\taskloss}{\mathcal{L}_{\mathrm{task}}}
\newcommand{\reconloss}{\mathcal{L}_{\mathrm{recon.}}}
\newcommand{\weightedloss}{\mathcal{L}_{\mathrm{weight}}}

\newcommand{\Aencoder}{A}
\newcommand{\Bdecoder}{B}

\newcommand{\Atilde}{\tilde{A}}
\newcommand{\Btilde}{\tilde{B}}

\newcommand{\KLQR}{K}
\newcommand{\taskoperand}{\KLQR x - \KLQR \Bdecoder \Aencoder x}

\newcommand{\reconoperand}{x - \Bdecoder \Aencoder x}

\newcommand{\thetatask}{\theta_{\mathrm{task}}}
\newcommand{\thetaencoder}{\theta_{\mathrm{enc.}}}
\newcommand{\thetadecoder}{\theta_{\mathrm{dec.}}}
\newcommand{\ftask}{f}
\newcommand{\xhat}{\hat{x}}
\newcommand{\zbottleneck}{Z}
\newcommand{\ninput}{n}
\newcommand{\moutput}{m}
\newcommand{\fencoder}{p}
\newcommand{\fdecoder}{q}

\newcommand{\trans}{\top}

\title{\LARGE \bf
Task-relevant Representation Learning for \\ Networked Robotic Perception}

\author{Manabu Nakanoya$^{*,1}$, Sandeep Chinchali$^{*,2}$, Alex Anemogiannis$^{3}$, Akul Datta$^{3}$, Sachin Katti$^{2}$, Marco Pavone$^{2}$
\thanks{* Equal Contribution. This work was supported in part by a NASA ULI grant on Safe Aviation Autonomy.}%
\thanks{$^{1}$ NEC Corporation, Tokyo, Japan {\small \tt{nakanoya@nec.com}}}
\thanks{$^{2}$ Stanford University, Stanford, CA, USA \tt{\{csandeep, skatti, pavone\}@stanford.edu}}
\thanks{$^{3}$ Unaffiliated {\small \tt{\{akuldatta,a.anemogiannis\}}@gmail.com}}
}

\begin{document}
\maketitle

\begin{abstract}
Today, even the most compute-and-power constrained robots \revision{can} measure complex, high data-rate video and LIDAR sensory streams. Often, \revision{such} robots, ranging from low-power drones to space and subterranean rovers, need to transmit high-bitrate sensory data to a remote compute server if they are uncertain or cannot scalably run complex perception or mapping tasks locally. 
However, today's representations for sensory data are mostly designed for \textit{human, not robotic}, perception and thus often waste precious compute or wireless network resources to transmit unimportant parts of a scene that are unnecessary for a high-level robotic task. This paper presents an algorithm to \revision{learn} \textit{task-relevant} representations of sensory data that are co-designed with \revision{a pre-trained robotic perception model's ultimate objective}. Our algorithm aggressively compresses robotic sensory data by up to \SC{11 $\times$} more than competing methods. Further, it \revision{achieves high accuracy and robust generalization} on diverse tasks including Mars terrain classification with low-power deep learning accelerators, neural motion planning, and \revision{environmental timeseries classification}.

\end{abstract}

\section{Introduction}
\label{sec:introduction}
Imagine a future Mars or subterranean rover that captures high-bitrate video and LIDAR sensory streams as it charts uncertain terrain, some of which it cannot classify locally. 
How should these robots represent, compress, and transmit their rich sensory data over bandwidth-limited wireless networks, especially if the intended audience is often a compute-intensive, remote machine learning model, not necessarily a human viewer? Indeed, even a single RGB-D (depth) camera stream produces upwards of 45 Megabytes/second of data \cite{nenci2014effective}, while the deep-space network only has a communication bandwidth of 0.5-4 Megabits/second \cite{DeepSpaceNetwork} \footnote{NASA estimate for Earth to Mars Reconnaissance Orbiter link \cite{DeepSpaceNetwork}.}.

More broadly, today's representations for sensory data mostly optimize for human, not robotic, perception and thus try to faithfully represent every pixel or point-cloud in a scene. Ideally, however, resource-constrained robots should represent only salient parts of sensory streams for remote perception and planning tasks to reduce the computational cost of encoding, storing, and transmitting sensory data. This paper presents a general algorithmic framework to \textit{learn} such concise, task-relevant representations. While recent work has used specialized deep neural networks (DNNs) to improve LIDAR or JPEG image compression \cite{tu2019point,emmons2019cracking,liu2018deepn,weber2019lossy,hotnets}, our key novelty is to use a general-purpose, pre-trained task module to \textit{guide} representation learning, which allows us to generalize to multiple sensor modalities.

\begin{figure}[t]
    \centering
    \includegraphics[width=1.0\columnwidth]{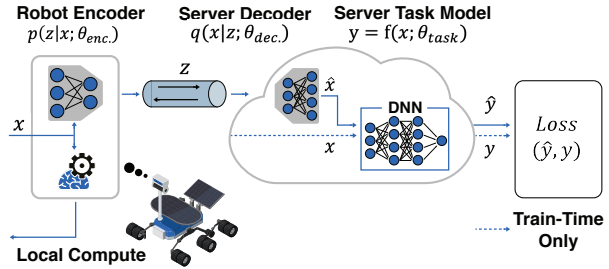}
    \caption{\textbf{Task-Relevant Communication for Perception}: 
    A compute-limited robot \textit{learns} how to compress sensory input $x$, transmit salient features $z$, and decode the input $\xhat$ so that it can directly leverage a pre-trained, potentially ``off-the-shelf'' task module $f(;\thetatask)$ at a central server.
By learning an encoder and decoder (gray) within the context of a pre-trained task module's goal, we only transmit minimal, salient information. Original sensory input $x$ and task output $y$ (dashed lines) are only used for training.}
    \label{fig:task_autoencoder}
\vspace{-1em}
\end{figure}

Specifically, as shown in Fig. \ref{fig:task_autoencoder}, our design provisions a pre-trained, differentiable task module at a central server, which allows multiple robots to share an upfront cost of training the model and benefit from centralized model updates. Our key technical insight is to \textit{co-design} a minimal sensory representation that is tightly coupled with the task module's objective. Henceforth, co-design means that the pre-trained task network parameters are fixed and the task objective guides what salient parts of a sensory stream to encode.
By also learning a task-relevant decoder, we enable robots to leverage a variety of pre-trained, publicly-available task modules without modifying their required input dimensions. Our approach is complementary to advances in on-robot computation and instead pertains to compute-intensive, potentially collaborative, tasks that require remote assistance.

\tu{Literature Review:}
Recent work has applied information bottleneck theory \cite{tishby2015deep} to build controllers that focus on actionable, task-relevant visual inputs for robust, generalizable navigation and grasping policies \cite{pacelli2020learning,pacelli2019task,sonar2020invariant}. \revision{In contrast, we introduce a novel algorithm for co-designing communication and machine perception, which uses pre-trained task modules to learn salient, efficiently-computable representations.} Our co-design approach differs from specialized solutions that use DNNs for JPEG image compression \cite{liu2018deepn,weber2019lossy} and instead uses diverse, off-the-shelf task modules to learn representations for images, obstacle point clouds, and environmental sensor timeseries.

Our work is also related to cloud robotics \cite{kehoe2015survey,kuffner2010cloud}, where robots use remote servers to augment their grasping \cite{li2018dex}, object recognition \cite{kehoe2013cloud}, and mapping capabilities \cite{mohanarajah2015cloud}. Prior work balances cloud computing accuracy with communication delay by learning when to query the cloud \cite{chinchali2019RSS,tanwani2020rilaas}. Rather than address \textit{when} to communicate, we instead address how to represent data for task-centric communication. \revision{Finally, our work is related to variational and multi-task autoencoders \cite{kingma2013auto, ghifary2015domain}, which compress inputs to minimize regularized reconstruction loss, often with domain-specific output layers in the multi-task case. In contrast, by instead focusing on task loss, we compress sensory data $11 \times$ more than standard autoencoders.}

\tu{Statement of Contributions:}
In light of prior work, our contributions are three-fold. First, we introduce a novel formulation that co-designs a distributed encoder/decoder with a fixed task network. Second, we develop a novel algorithm that aggressively compresses sensory data for diverse robotic tasks. \revision{Third, we show how to flexibly allocate computation between a compute-limited robot and server, which renders our approach compatible with low-power DNN accelerators like the Google Edge Tensor Processing Unit (TPU) \cite{edgeTPUwebsite}.}

\tu{Organization:}
This paper is organized as follows. In Sec. \ref{sec:problem_statement}, we introduce a novel, general problem of co-designing
sensory representations for robotic perception. To address this problem, Sec. \ref{sec:task_specific_algorithm} presents our co-design algorithm and \revision{highlights significant compression gains for an illustrative example of linear systems}.
Then, Sec. \ref{sec:experiments} evaluates our algorithm on diverse perception tasks with compute-efficient DNNs. Finally, Sec. \ref{sec:conclusion} concludes with future directions.

\section{Problem Statement}
\label{sec:problem_statement}
\revision{We now introduce the core compute modules for information flow between a robot and a central server (Fig. \ref{fig:task_autoencoder}), and then formalize our problem statement}. \revision{First, a robot measures a sensory input $x \in \reals^n$, such as an image or LIDAR point cloud. Without loss of generality, $x$ could represent a single sensory sample a robot wishes to transmit or a window $w$ of correlated samples measured from time $t-w$ to $t$, such as a segment of video, denoted by $\mathbf{x}= x^{t-w:t}$.} 

\noindent \tu{Robot Encoder:}
The encoder maps raw input $x$ to a concise representation  $z \in \reals^\zbottleneck$, denoted by $z \sim  \fencoder(z | x; \thetaencoder)$, \revision{where $\thetaencoder$ are encoder model parameters, such as learned DNN weights}. Henceforth, $z$ is referred to as a bottleneck representation, since it is compressed via an information bottleneck, such as a DNN hidden layer, to a size $\zbottleneck \ll n$. 

\noindent \tu{Server Decoder:}
The encoder transmits bottleneck representation $z$ over a wireless link to a differentiable decoder at a central server, which generates a reconstructed estimate of the raw sensory input $\xhat \sim \fdecoder(x | z; \thetadecoder)$. \revision{The parameters of a decoder model, such as a DNN, are denoted by $\thetadecoder$}. 

\noindent \tu{Server Task Network:}
Finally, the decoded input $\xhat$ is \textit{directly passed} through a pre-trained, differentiable task module, which serves as the key distinction of our work from a standard autoencoder. \revision{An example} of a task module could be a DNN object detector that predicts object locations and classes $\yhat = \ftask(\xhat; \thetatask)$ using model parameters $\thetatask$. Crucially, by decoding to the original input dimension $n$, our approach enables robots to directly utilize a plethora of pre-trained, publicly-available task modules that expect an input dimension $n$ without re-training on custom bottleneck representations $z$.
In practice, the encoder, decoder, and task module can all be DNNs. We assume sensory input $x$ and its label $y$ are drawn from a domain-specific distribution $\mathcal{D}$, such as the space of all Mars terrain images and labels. 

\subsection{Task and Reconstruction Optimization Objectives}
Our principal objective is to minimize the {\em task loss} $\taskloss(y, \yhat; \thetatask)$, which compares the resultant task outputs $y = f(x; \thetatask)$  using original input $x$ and predicted outputs $\yhat = \ftask(\xhat; \thetatask)$ using a decoded input $\xhat$. Our key technical insight is that robotic perception tasks can often tolerate considerable distortion in the input estimate $\xhat$, such as omitting irrelevant parts of an image/map for classification/planning, as long as the downstream task module $\ftask(; \thetatask)$ can still achieve its goal. \revision{Thus, decoded input $\xhat$ should only represent task-relevant features, which enables highly-compressed representations $z$ to be transmitted over a wireless link}.

Optionally, a roboticist might want to also minimize reconstruction loss, in cases where decoded inputs $\xhat$ should be moderately human-interpretable to view or debug portions of an image. For such scenarios, loss function $\reconloss(x, \xhat; \thetaencoder, \thetadecoder)$ incentivizes faithful reconstruction of a scene, which, in practice, could be the standard variational autoencoder regularized loss \cite{kingma2013auto}. We provide a general, flexible co-design framework that allows a roboticist to optimize a weighted combination of task and reconstruction loss, with reconstruction loss weight $\lambda \ge 0$: 
\vspace{-1em}

\begin{small}
\begin{align}
    \weightedloss(x,\xhat, y, \yhat; \thetatask, \thetaencoder, \thetadecoder) = \nonumber \\ \taskloss (y, \yhat; \thetatask) + \lambda \reconloss (x, \xhat; \thetaencoder, \thetadecoder). 
\label{eq:weighted_loss}
\end{align}
\end{small}
Our experiments evaluate the scenario of strictly optimizing for task loss ($\lambda=0$) as well as various values of $\lambda > 0$, which introduces a regularization term to incentivize highly-compressed representations that still yield human-interpretable reconstructions. Having defined our optimization objective, we now formalize our problem statement. 

\subsection{Problem Statement}
\begin{problem}[Sensory Co-design for Machine Perception]
\label{prob:codesign}
Given a differentiable task module $f(; \thetatask)$ with fixed, pre-trained parameters $\thetatask$, fixed bottleneck dimension $\zbottleneck$, and reconstruction loss weight $\lambda \geq 0$, find robot encoder and server decoder parameters $\thetaencoder$ and $\thetadecoder$ that minimize weighted loss (Eq. \ref{eq:weighted_loss}) over data distribution $\mathcal{D}$:\vspace{-1em}

    \begin{small}
    \begin{align*}
        \thetaencoder^{*}, \thetadecoder^{*} = \argmin_{\thetaencoder, \thetadecoder} \mathop{\mathbb{E}}_{(x,y)\sim \mathcal{D}} \weightedloss (x, \xhat, y, \yhat; \thetatask, \thetaencoder, \thetadecoder), 
    \end{align*}
    \end{small}
    where $\xhat = \fdecoder(\fencoder(x; \thetaencoder); \thetadecoder)$ and $\yhat = \ftask(\xhat;\thetatask)$.
\end{problem}

\revision{Prob. \ref{prob:codesign} is widely applicable to resource-limited robots, such as space and mining rovers, that can only unlock large benefits of remote computation if they send \textit{task-relevant} information over a bandwidth-limited network. The key novelty of our formulation is that we leverage knowledge of a task objective, through $\ftask(;\thetatask)$, to guide task-relevant representation learning of encoder/decoder parameters $\thetaencoder, \thetadecoder$. Further, we enable a roboticist to set bottleneck size $\zbottleneck$ in a principled, precise manner based on a network's maximum allowable data-rate. For example, the data-rate could be the size of $Z$ 32-bit floating point values sent at a certain communication frequency.}

\subsection{Illustrative Example: Linear Systems Setting}
\label{subsec:linear_setup}
To illustrate the compression benefits of optimizing for task loss in Prob. \ref{prob:codesign}, we consider a toy example where the encoder, decoder, and task module are matrices.
\revision{Specifically, consider a simple robotic sensor network where a central server must estimate a function $y = \KLQR x$ of a potentially large sensor measurement $x$, \textit{without} necessarily sending the full input $x$ over a wireless link.}
First, robot encoder matrix $\Aencoder = \thetaencoder \in \reals^{\zbottleneck \times \ninput}$ generates encoding $z = \fencoder(x; \thetaencoder) = \Aencoder x$. After $z$ is sent over a wireless link, it is decoded to $\xhat = \fdecoder(z; \thetadecoder) = \Bdecoder z$ by linear decoder $\Bdecoder = \thetadecoder \in \reals^{\ninput \times \zbottleneck}$. Finally, task matrix $\KLQR \in \reals^{\moutput \times \ninput}$ generates the output of the linear estimation problem $y \in \reals^\moutput$. 
\revision{The task loss penalizes error in the linear estimate $y=\KLQR x$, and the reconstruction loss optionally penalizes error in the sensor estimate $\xhat$ in case elements of $\xhat$ need to be sanity-checked. Both task and reconstruction loss are quadratic, yielding weighted loss:}
\begin{align}
    \weightedloss = \underbrace{\Vert (\KLQR x - \KLQR \Bdecoder \Aencoder x) \Vert^2_2}_{\mathrm{task~loss}} + \lambda \underbrace{\Vert(\reconoperand)\Vert^2_2}_{\mathrm{recon. loss}}.
    \label{eq:weighted_loss_linear}
\end{align}
To show the full benefits of task-based compression, we now provide an analytical solution for Prob. \ref{prob:codesign} for the special case when $\lambda=0$ (pure task loss objective).
\noindent \begin{thm}[Linear Task-Aware Compression]
Consider task matrix $\KLQR \in \reals^{\moutput \times \ninput}$ with rank $r$ and compact singular value decomposition (SVD) $K = U \Sigma V^{\trans}$, where $U\in\reals^{\moutput \times r}$ and $V^{\trans} \in \reals^{r \times n}$ are semi-unitary.
Then, for bottleneck dimension $r$, setting $\Aencoder = V^{\trans}$ and $\Bdecoder = V$ solves Prob. \ref{prob:codesign} with zero task loss $\Vert (\KLQR x - \KLQR \Bdecoder \Aencoder x) \Vert^2_2$ (Eq. \ref{eq:weighted_loss_linear} with $\lambda=0$) for any $x \in \reals^{n}$. 
Further, there are no encoder and decoder matrices with bottleneck $Z < r$ that achieve zero task loss.
\end{thm}
\noindent \begin{proof}
The task loss is zero when, for any $x \in \reals^{\ninput}$, $y = \KLQR x = \yhat = \KLQR \Bdecoder \Aencoder x$. This is achieved when $\Bdecoder = V$ and $\Aencoder = V^{\trans}$:
\begin{small}
\begin{align*}
    \yhat = \KLQR \Bdecoder \Aencoder x &= \underbrace{(U \Sigma V^{\trans})}_{SVD(K)} \times \underbrace{V}_{\Bdecoder} \times \underbrace{V^{\trans}}_{\Aencoder} \times x \\
							&= U \Sigma V^{\trans} x = Kx = y,  
\end{align*}
\end{small}
\vspace{-1em}

\noindent where $V^{\trans}V = I_{r \times r}$ since $V$ is semi-unitary. We now show there are no other solutions with bottleneck $Z< r$. For the sake of contradiction, suppose there exist $\Atilde\in\reals^{Z\times n}, \Btilde \in \reals^{n\times Z}$ with $Z < r$ that achieve zero task loss. Then:
\begin{small}
\begin{align*}
    \text{rank}(\Btilde\Atilde) \leq ~&\text{min}(\text{rank}(\Atilde), \text{rank}(\Btilde)) \leq Z, \text{and} \\ 
    \text{rank}(\KLQR\Btilde\Atilde) \leq ~&\text{min}(\text{rank}(K), \text{rank}(\Btilde\Atilde)) \leq Z.
\end{align*}
\end{small}
However, in order for $\Atilde, \Btilde$ to achieve zero task loss, we must have $Kx=K\Btilde\Atilde x$ for any $x \in \reals^{\ninput}$, meaning we must have $K = K\Btilde\Atilde$. However, this is a contradiction since $\text{rank}(K\Btilde\Atilde)\leq Z < r=\text{rank}(K)$. Thus, we conclude there are no solutions that achieve zero task loss for $Z < r$.
\end{proof}

\tu{Compression benefits:}
Our key insight is that, by co-designing with fixed task matrix $\KLQR$, we achieve zero task loss with a compression gain of $\frac{n}{\text{rank}(K)}$ compared to sending full input $x$. However, we do not necessarily have zero \textit{reconstruction loss} since matrix $V^{\trans}$ is only semi-unitary, meaning it is not required for $V V^{\trans} = \Bdecoder \Aencoder$ to be an identity matrix. \SC{In fact, if we want zero reconstruction loss, then we have $\hat{x} = \Bdecoder \Aencoder x= x$ only when $\Bdecoder = \Aencoder^{-1}$, requiring $\Bdecoder$ and $\Aencoder$ to be square $n \times n$ matrices, yielding no compression gain.} Crucially, our solution differs from na\"ively just computing $y=Kx$ directly at the robot, since we only transmit a small representation $z=V^{\trans}x$ and use the modular task matrix $\KLQR$ at the server. Finally, we note that the choice of $A=V^{\trans}$ and $B=V$ is not unique for bottleneck $r$. For example, we can scale the encoder to $aA$ and decoder to $bB$ where $ab=1$.

\begin{figure}
    \centering
    \subfloat{
    \includegraphics[width=0.5\columnwidth]{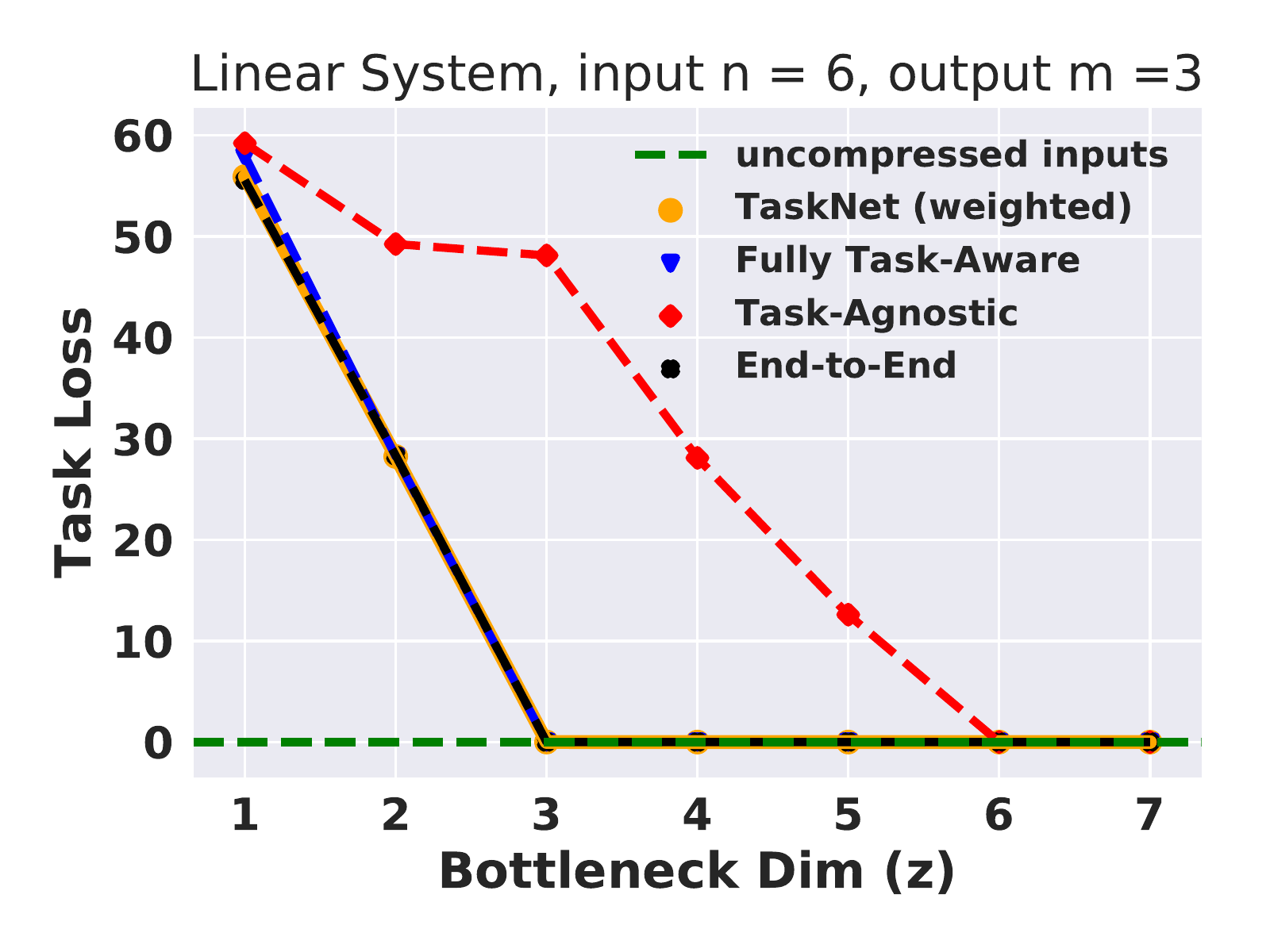} 
    \label{fig:LQR_3D_task_loss}
    }
    \subfloat{
        \includegraphics[width=0.5\columnwidth]{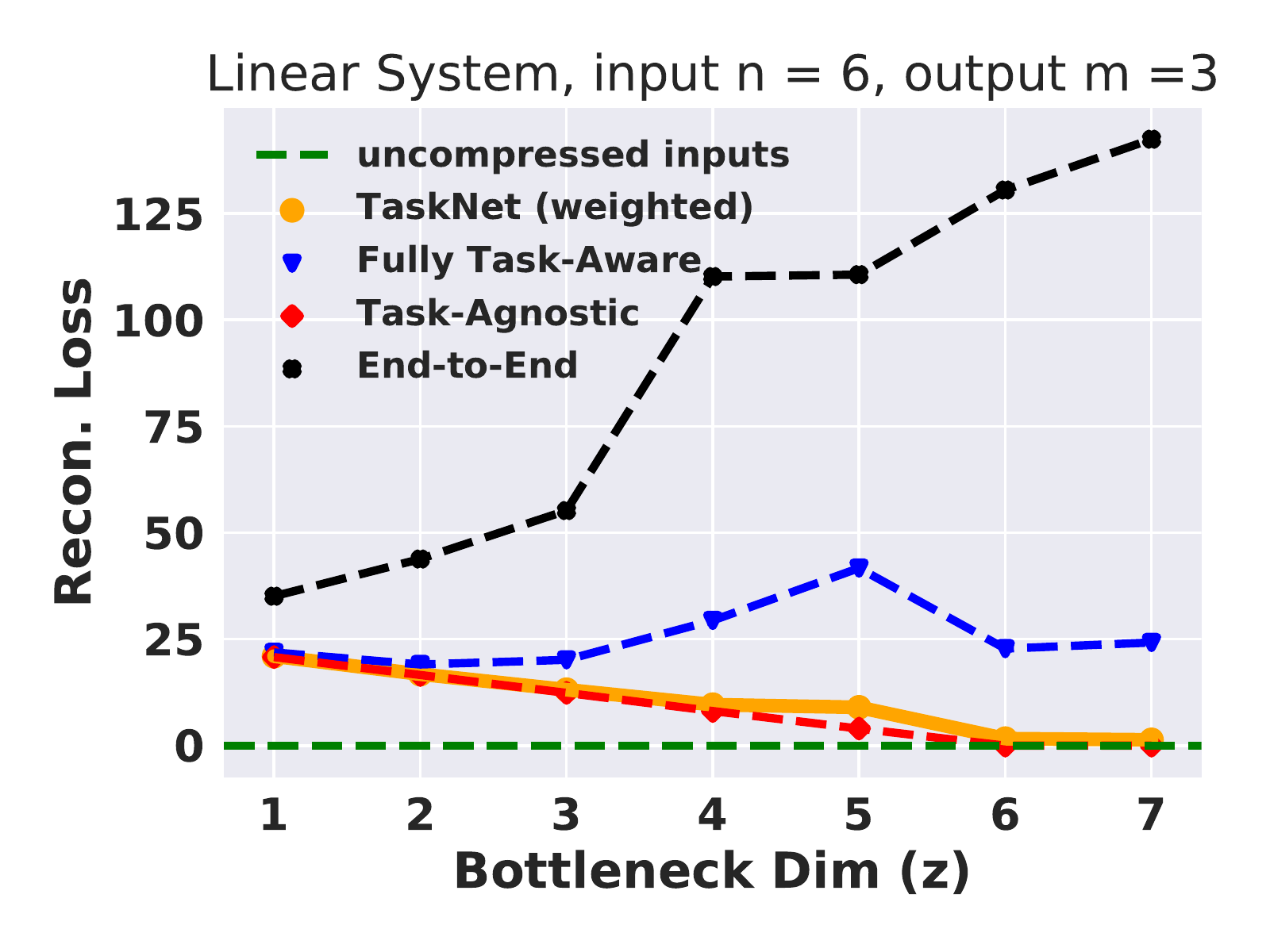} 
    \label{fig:LQR_3D_recon_loss}
    }
    \caption{\textbf{Task-aware compression for linear systems.}  When the encoder, decoder, and task module $K$ are matrices, we achieve zero task loss by only sending a representation of size $\mathrm{rank}(K)$ compared to a measurement of size $n$. Our co-design scheme (orange, blue) yields lower task loss for much smaller $z$ compared to task-agnostic methods (red). Further, we achieve better reconstruction loss and more stable training than an end-to-end method (black), which does not guide learning by fixing task module $K$.}
    \label{fig:LQR_simulation_3D}
\vspace{-1em}
\end{figure}

\tu{Weighted linear case:}
For the weighted $\lambda > 0$ case, we analytically compute the gradient of weighted loss (Eq. \ref{eq:weighted_loss_linear}) and solve Prob. \ref{prob:codesign} using gradient descent. Fig. \ref{fig:LQR_simulation_3D} shows results for a toy linear problem with input and output dimensions $n=6$ and $m=3$, where we achieve optimal task performance with a $2\times$ compression gain. Due to space constraints, we provide further details online \footnote{Our technical report \cite{tasknet} and Edge TPU software are available at \\ \url{https://sites.google.com/view/tasknet}.}. 
\revision{Our toy example shows that, even for linear systems, we achieve large compression gains by co-designing with a task objective, which we extend to complex DNNs in the next section.}

\section{Task-Aware Co-design Algorithm}
\label{sec:task_specific_algorithm}
In general, analytically solving Problem \ref{prob:codesign} is challenging, especially when the task network and encoder/decoder are DNNs. We now present Algorithm \ref{alg:train}, which learns an encoder/decoder that approximately solves Problem \ref{prob:codesign}. Importantly, our algorithm yields experimental results that match  analytical calculations for simple linear settings and also scales to deep learning tasks, presented in Sec. \ref{sec:experiments}.

Alg. \ref{alg:train} takes as input a bottleneck dimension $\zbottleneck$, set based on communication data-rate limits, and user-desired reconstruction weight $\lambda$, on line 1. Then, it randomly initializes the encoder/decoder parameters on line 2, which are learned during training \textit{unlike} task parameters $\thetatask$ (line 3). The main loop repeats for $T$ learning rounds, where we sample inputs from labeled training dataset $\mathbf{D} = \{x^i, y^i\}$ where each $(x^i, y^i) \sim \mathcal{D}$ (line 5), encode/decode sensory data (lines 6-7), and finally invoke the task module on line 8. The key step for co-design is line 9, where we backpropagate weighted loss $\weightedloss$ (Eq. \ref{eq:weighted_loss}) to update parameters $\thetaencoder, \thetadecoder$, while keeping the task parameters $\thetatask$ \textit{fixed}. \revision{Importantly, Alg. \ref{alg:train} can be run \textit{offline} given a training dataset and desired bottleneck $Z$ to avoid passing large gradients over a wireless link. Then, a robot can deploy the trained encoder and decoder, and periodically improve them with more field data.}

\begin{small}
\begin{algorithm}[h]
\DontPrintSemicolon
\SetNoFillComment 
\SetAlgoLined
 Set bottleneck dimension $\zbottleneck$, reconstruction weight $\lambda$\;
 Randomly initialize encoder/decoder params. $\thetaencoder^{0}$, $\thetadecoder^{0}$\;
 Clamp pre-trained model $f(;\thetatask)$ params. $\thetatask$ \; 
 \For{$\tau~ \gets0$ \KwTo $\Nepochs$}{
    Sample $\{x,y\}$ from dataset $\mathbf{D}$ with $y=\ftask(x;\thetatask)$ \;
    Encode $z \sim p(z ~|~ x; \thetaencoder^{\tau})$ \;
    Decode $\xhat \sim q(x ~|~ z; \thetadecoder^{\tau})$ \;
    Compute Predictions $\yhat = f(\xhat; \thetatask)$ \;
    $\thetaencoder^{\tau + 1}, \thetadecoder^{\tau + 1} \gets \textsc{BackProp}\big[\weightedloss(x, \xhat, y, \yhat; ~\thetatask,\thetaencoder^{\tau},\thetadecoder^{\tau}) \big]$ \; 
 }
 \KwResult{Return learned parameters $\thetaencoder^{T}, \thetadecoder^{T}$}
 \caption{Task-Relevant Compression Co-design}
 \label{alg:train}
\end{algorithm}
\end{small}

\subsection{Wide applicability of our Co-design Algorithm}
\revision{While simple, Alg. \ref{alg:train} is a powerful, general solution for task-relevant communication. 
A principal benefit is that we can either invoke a pre-trained task module fully remotely, or also flexibly adjust the split of computation between a resource-constrained robot and server. 
In the latter scenario, we can take a large, pre-trained DNN and only execute a fraction of layers locally on the robot to flexibly adjust for resource constraints. Then, we can insert a minimal, learned encoder $\fencoder$ to compress an intermediate result, transmit bottleneck $z$, and then decode with $\fdecoder$ to continue computation with the bulk of the pre-trained model at a server. Importantly, \textit{only} the distributed encoder and decoder are trained, unlike the fixed task module. Further, by adjusting reconstruction weight $\lambda$, which serves as a regularizer, we can significantly reduce reconstruction error with only a marginal gain in bottleneck size $z$. Our ability to flexibly set $\lambda$ leads to the following variants of Alg. \ref{alg:train}}:

\noindent 1. \tu{Task-Aware-Weighted (\name)}: Our core contribution is the weighted task-aware training scheme, henceforth referred to as \name. 
By accounting for the pre-trained task network $\ftask(;\thetatask)$ when encoding sensory inputs, \name achieves low task loss for much less transmitted data $z$. 
\name's reconstruction loss weight $\lambda$ (Eq. \ref{eq:weighted_loss}) can be flexibly set by a roboticist per application. All our experiments performed a scan over evenly-spaced values of $\lambda$ and only plotted one representative curve, in orange, for visual clarity.

\noindent 2. \tu{Fully Task-Aware}: To quantify the full compression benefits of optimizing only for an end-task, we consider the special case of \name when $\lambda = 0$, which is suitable for scenarios when video inputs are automatically classified in real-time by a DNN and not intended for a human viewer. This scheme, always colored in blue, achieves the lowest task loss for the smallest representation size $\zbottleneck$, which is ideal for scenarios with strict network bandwidth limits. 
\revision{We now evaluate both variants of Alg. \ref{alg:train}, as well as benchmark schemes, on large-scale deep learning experiments.}

\section{Experimental Results}
\label{sec:experiments}
\revision{To evaluate \name, we first introduce two benchmark algorithms and then describe common evaluation metrics for all schemes. Then, we highlight the wide utility of \name across various sensing modalities and task DNN modules, for tasks ranging from Martian terrain classification to motion planning and environmental timeseries classification. Further, we benchmark performance on the standard MNIST dataset.}

\subsection{Evaluation Metrics and Benchmark Algorithms}
\revision{The principal objective of Prob. \ref{prob:codesign} is to minimize task loss, and optionally minimize weighted reconstruction loss, in scenarios where decoded inputs $\xhat$ need to be inspected. As such, we show that \name achieves (A) low task loss (i.e. high task accuracy) and (B) flexibly achieves low reconstruction loss for small bottleneck sizes $z$. To assess the effects of compression on task loss, we quantify the task loss achieved by passing uncompressed, \textit{original} sensory inputs $x$, without an information bottleneck, into the task network $\ftask$. This metric, henceforth referred to as the \textit{uncompressed input task loss}, represents the best task accuracy we can achieve without network constraints, and is pictured by the green dashed line in all task loss figures (e.g. Fig. \ref{fig:MNIST_task_recon_loss}). Further, all figures show results on each domain's \textit{test} dataset.
We test the above metrics on the following two benchmarks that represent conventional paradigms for networked perception:}

\noindent 1. \tu{Task-Agnostic}: 
For each bottleneck dimension $\zbottleneck$, this scheme trains a standard variational autoencoder to minimize reconstruction loss, and simply passes the decoded input $\xhat$ through the pre-trained task module $\ftask(;\thetatask)$. The task-agnostic scheme is always colored in red in all figures. 

\noindent 2. \tu{End-to-End}: This scheme has the exact same encoder, decoder, and task module architectures as the others, but allows the task parameters $\thetatask$ to be updated. It takes longer to train and performs worse than \name for small bottlenecks $z$ when the networks get stuck in local minima during training, such as in Fig. \ref{fig:MNIST_task_recon_loss}. 
In essence, it does not exploit the structure of task network $\ftask$ for faster, more stable representation learning. \revision{We now evaluate all schemes.}

\subsection{Digit Classification with an Information Bottleneck}
\label{subsec:MNIST}
\begin{figure}
    \centering
    \subfloat{
        \includegraphics[width=0.5\columnwidth]{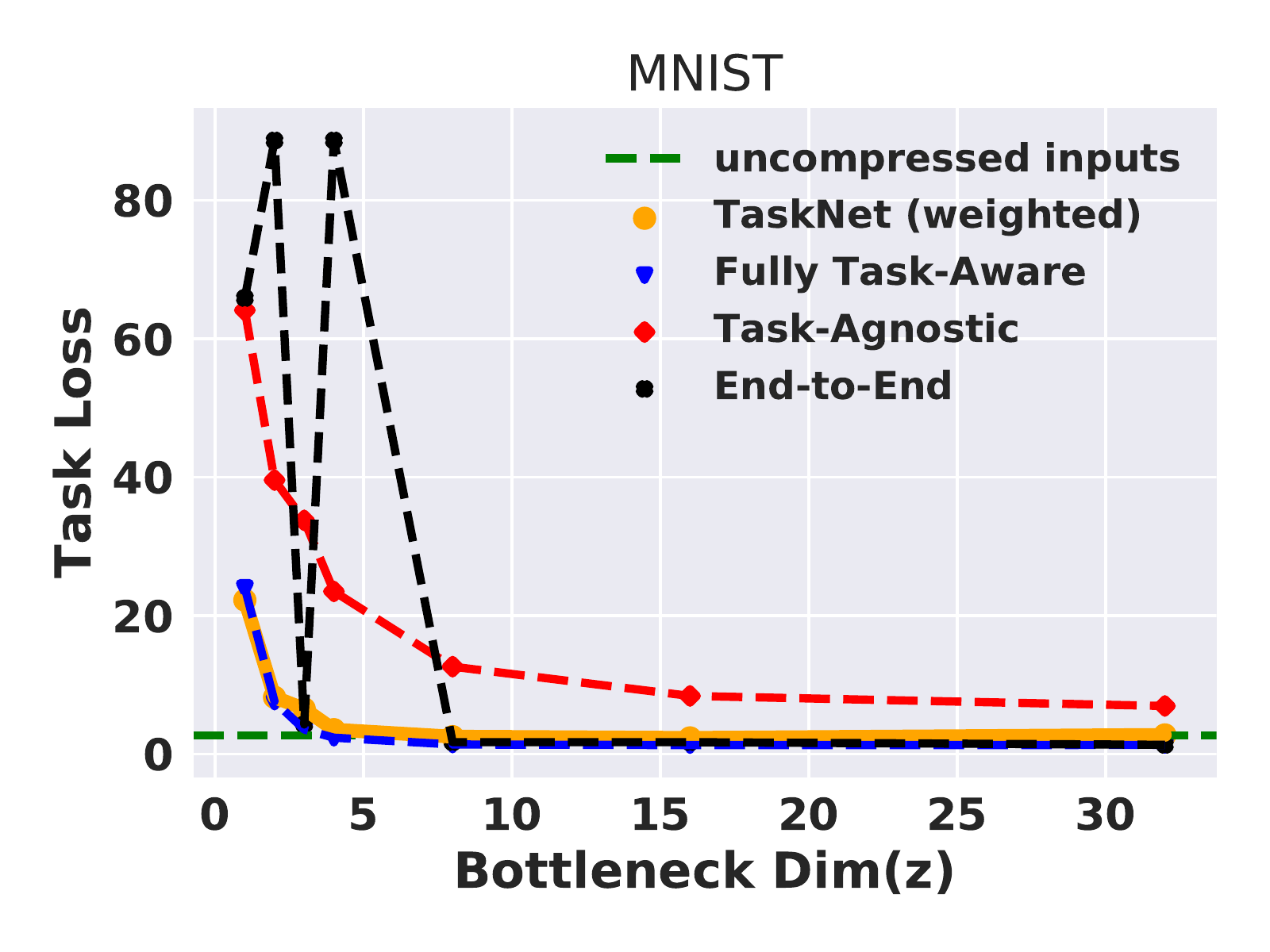}
        \label{fig:MNIST_task_loss}
    }
    \subfloat
    {
        \includegraphics[width=0.5\columnwidth]{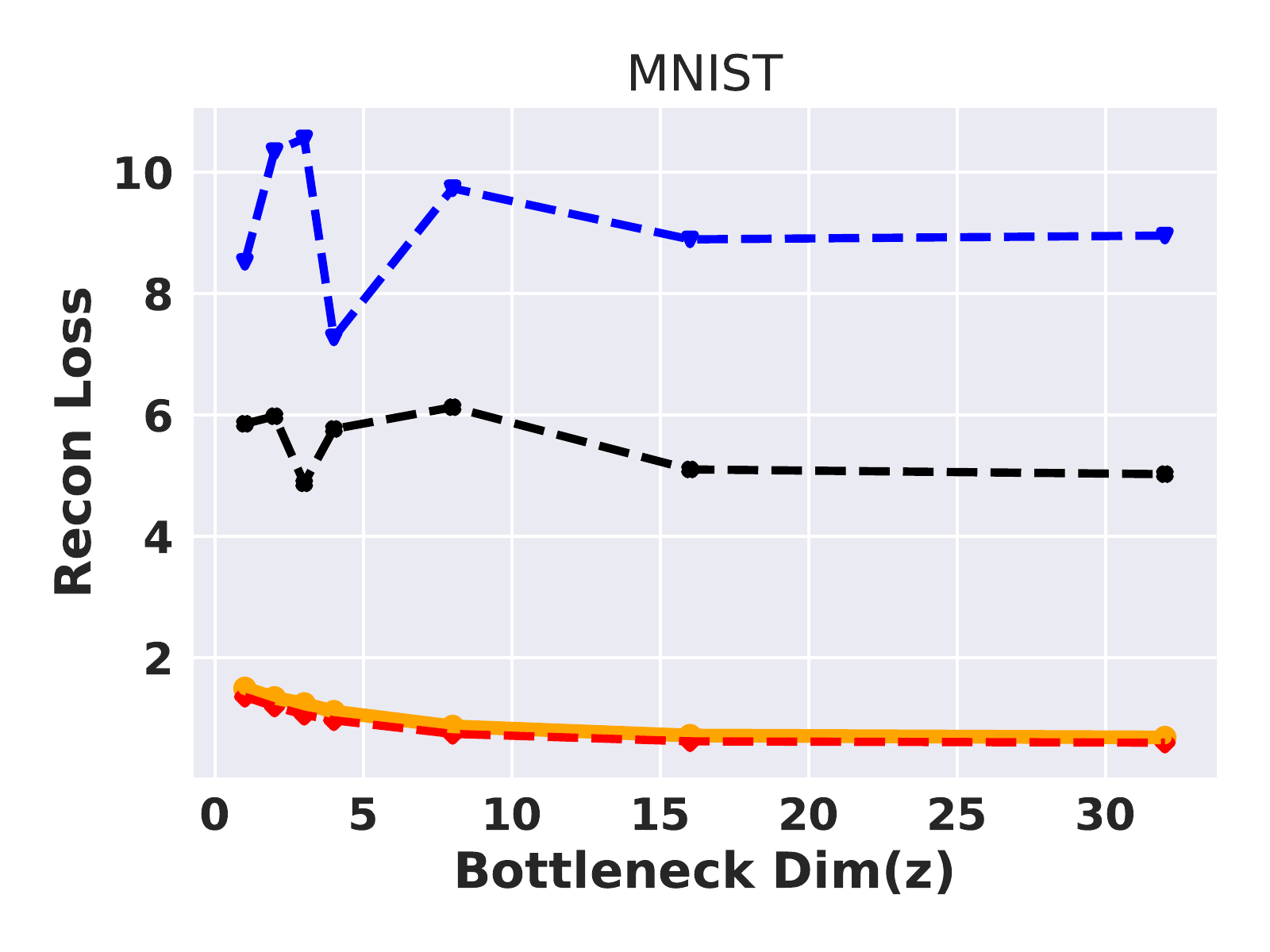}
        \label{fig:MNIST_recon_loss}
    }
    \caption{\textbf{MNIST test dataset results:}
        \name achieves the same lower-bound classification error as passing \textit{original} images through the task network (green) for a small bottleneck $\zbottleneck=3$. In contrast, a task-agnostic method, in red, obtains worse classification error for even larger $\zbottleneck=32$ since it does not emphasize salient features for classification. 
    }
   \label{fig:MNIST_task_recon_loss}
\vspace{-1em}
\end{figure}

We first benchmark \name on the standard MNIST \cite{lecun1998mnist} dataset, where task network $\ftask(;\thetatask)$ was a publicly-available ResNet-50 DNN for digit classification \cite{robustness}. Fig. \ref{fig:MNIST_task_loss} illustrates that \name (orange) achieves the same \revision{lower-bound classification error} as passing \textit{original, unperturbed} images to the task DNN, even for images that are compressed with an extremely small bottleneck of size $\zbottleneck = 3$. In contrast, the task-agnostic scheme (red) is still not able to achieve the \revision{lower bound classification error} even for a much larger bottleneck of $\zbottleneck = 32$, showing that \name improves the compression ratio by $10\times$ over a standard autoencoder with \textit{better} task accuracy. Fig. \ref{fig:MNIST_recon_loss} shows that \name decodes images $\xhat$ with low reconstruction loss, which yields human-interpretable, but highly compressible, decoded samples that are visualized in our online report \cite{tasknet}. In the end-to-end scheme, the task network is also learned during training.

\subsection{Compute-Efficient Mars Terrain Classification}
\label{subsec:mars}
To test \name with complex vision DNNs, we consider a scenario where a low-power Mars rover encodes terrain images for remote classification at a powerful orbiting compute server, a scenario inspired by recent proposals \cite{KISS}. We used Martian terrain images from NASA's HiRise Dataset \cite{hirise}, consisting of 8 classes \revision{including dunes, craters, impact ejecta, and other formations}. 
Task network $\ftask(;\thetatask)$ is an EfficientNet-B0 DNN, which is the smallest of a family of DNNs that trade off accuracy and model complexity \cite{tan2019efficientnet}.
Our EfficientNet has \SC{5.36 million} parameters and classification accuracy of 95.7\% on \SC{7303} HiRise test images.

Given the relatively large size of the vision DNN, we want to flexibly split computation between a compute-limited robot and server. Notably, we can exploit a common architecture in many modern DNNs, which are composed of \textit{sequential blocks} of several convolutional layers, to adjust the extent of on-robot compute. For example, EfficientNet block-4c consists of all parameters from the input to a specific layer $l$ and block-5c has all successive parameters until layer $l'>l$. As per Alg. \ref{alg:train}, we first train the full, modular EfficientNet. Then, we constrain the robot to only run part of the large DNN until a specific block $b$, which yields an intermediate convolutional feature map, and then insert our minimal task encoder $\fencoder(;\thetaencoder)$ to compress the feature map to encoding $z$. Encoding $z$ is transmitted to a remote server, which uses our task decoder $\fdecoder(;\thetadecoder)$ to reconstruct the original feature map generated by block $b$ and resumes computation for a final classification. Importantly, only the encoder/decoder are learned with the \textit{fixed} EfficientNet.

\begin{figure}
    \centering
    \subfloat{
    \includegraphics[width=0.49\columnwidth]{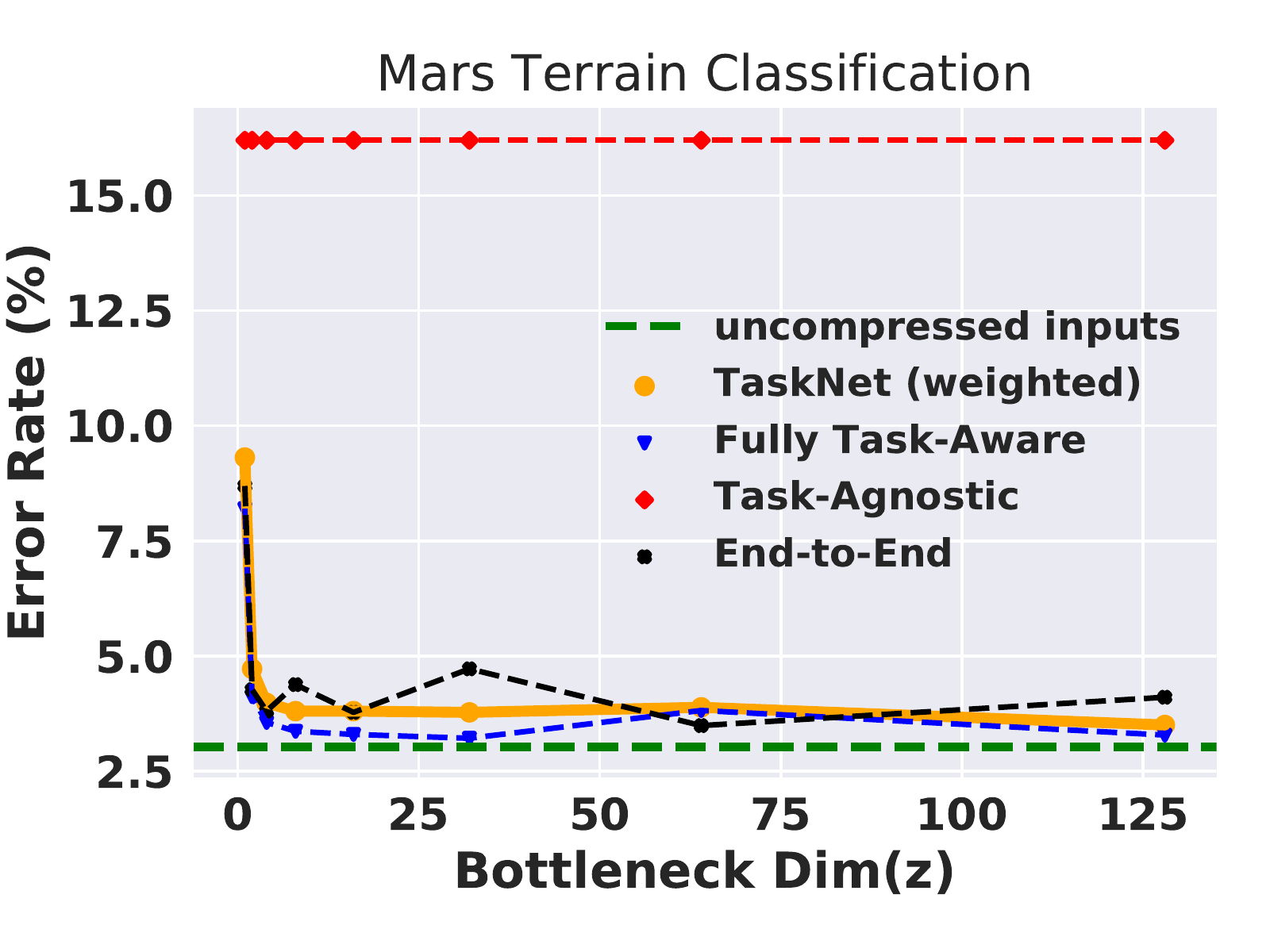}
    \label{fig:mars_task_loss}
    }
    \subfloat{
    \includegraphics[width=0.49\columnwidth]{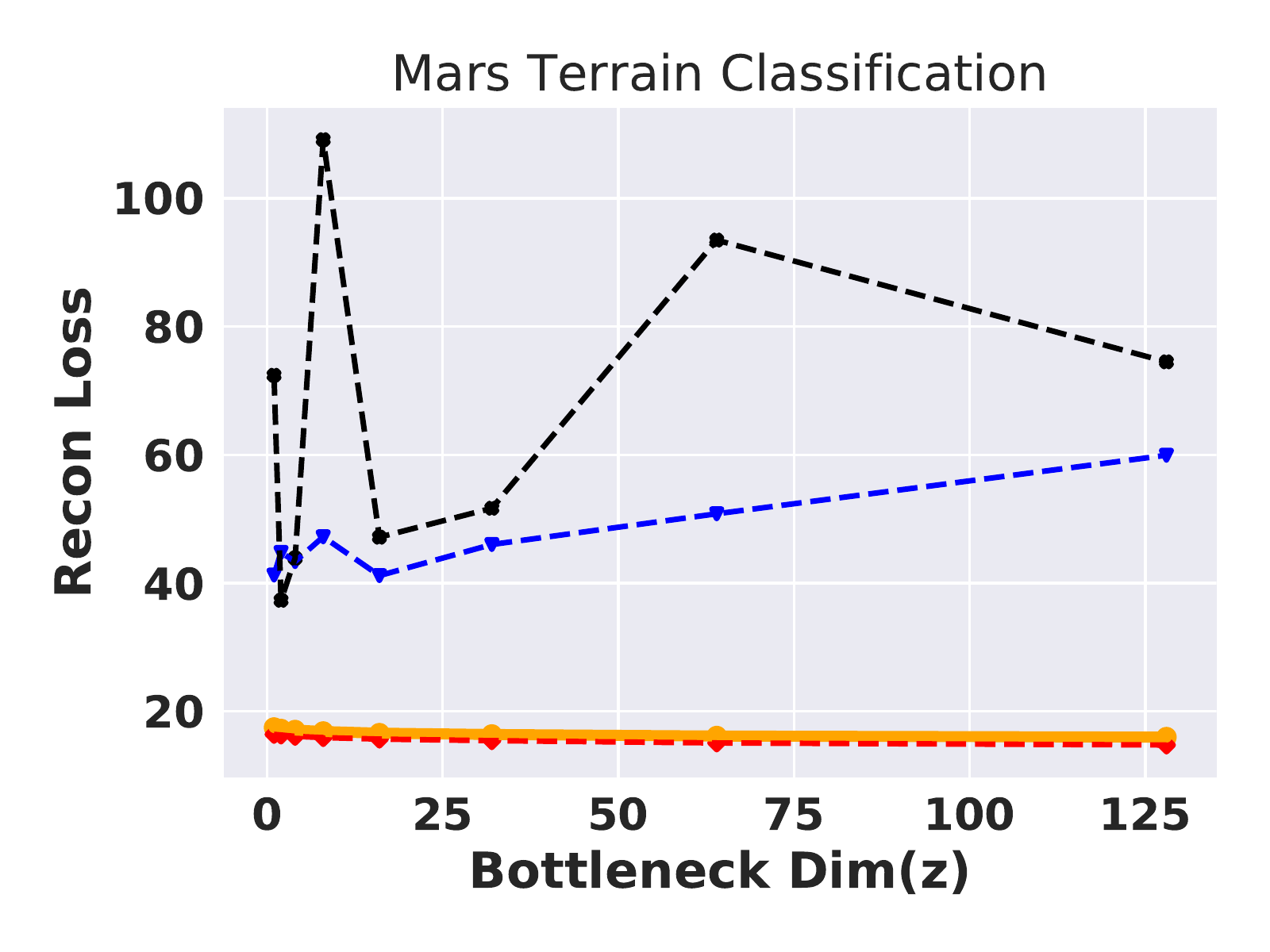}
    \label{fig:mars_recon_loss}
    }
    \caption{\textbf{Terrain classification.} \name achieves much lower classification error than task-agnostic methods (red), where even a bottleneck dimension of $z=128$ corresponds to sending only \SC{$0.34$ \%} of data compared to original images.}
    \label{fig:mars_hirise_specific_split}
\end{figure}

Our extended report \cite{tasknet} details several configurations where the robot only encodes inputs up to EfficientNet block $b$, but still achieves within 1.5\% of the \lossname. 
For example, Fig. \ref{fig:mars_hirise_specific_split} shows by computing until block-5c, \name 
achieves \SC{3.87 $\times$} lower classification error than conventional task-agnostic methods, with a bottleneck as small as $\zbottleneck = 16$. This enables our scheme to send only \SC{$<0.34\%$} of data compared to the original image. 

\begin{figure}
    \centering
    \subfloat[Martian Terrain Classification]{
    \includegraphics[width=0.49\columnwidth]{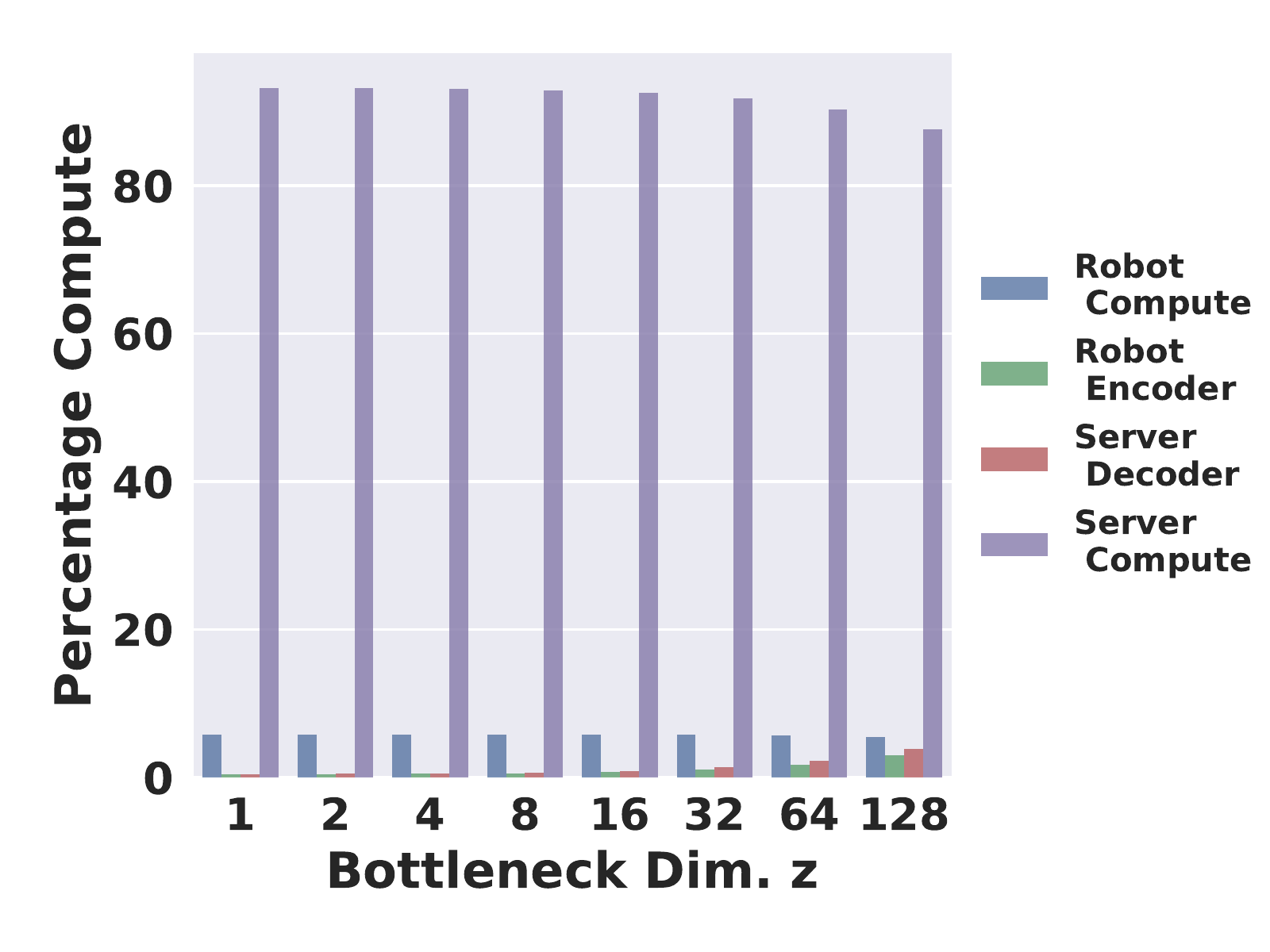}
    \label{fig:efnet_size}
    }
    \subfloat[Neural Motion Planning]{
    \includegraphics[width=0.49\columnwidth]{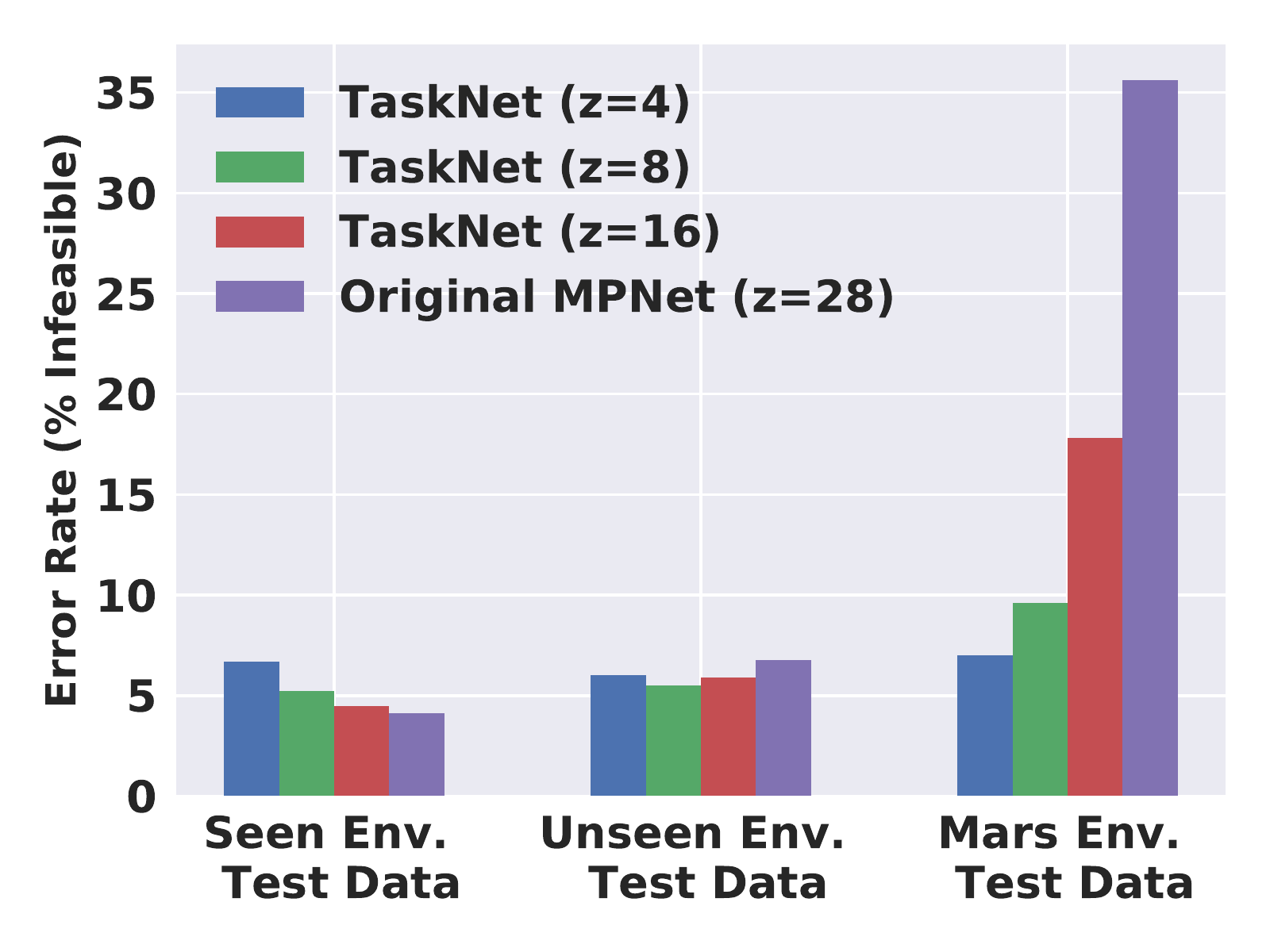} 
    \label{fig:mars_generalization}
    }
    \caption{(a) By only encoding task-relevant information, \name reduces on-robot computation compared to what is delegated to a remote server (purple). (b) \name generalizes better than the original \MPNET on out-of-domain Mars terrain data to find more feasible paths (y-axis).} 
    \label{fig:mars_barplots}
    \vspace{-1em}
\end{figure}

\textbf{Robot compute savings}:
Notably, by only devoting computational effort to encode task-relevant information, \name reduces robot computation to only \SC{8.5\%} of overall compute, compared to the \SC{91.5 \%} delegated to a remote server (Fig. \ref{fig:efnet_size}). 
As such, \name is significantly more efficient than na\"ively just splitting a compute-intensive DNN in half or running it fully on-board, which could be infeasible for micro-robots or miniature satellites like KickSats \cite{manchester2013kicksat}.

\subsection{Robotic Motion Planning with Neural Networks}
\label{subsec:motion_planning}

\revision{To demonstrate \name's utility for motion planning, we consider a scenario where a low-power drone takes an aerial terrain image and sends a compressed representation to a remote server, which generates a collision-free motion plan for a swarm of ground robots.} Our task network $\ftask(;\thetatask)$ is a \textit{Motion Planning Network} (\MPNET), a recently published planner \cite{qureshi2019motion} that generates collision-free paths an order of magnitude faster than conventional sampling-based planners such as RRT* \cite{karaman2011anytime}, but with virtually the same path cost. \MPNET consists of an obstacle encoder network, such as a contractive autoencoder, which maps an obstacle point-cloud to an embedding of dimension up to $\zbottleneck = 28$. Then, \MPNET's planning network maps the obstacle embedding $z$ as well as current and goal robot configurations to predict the next navigation waypoint. The original \MPNET DNN planners for two and three dimensional environments are up to \SC{41 MB} and \SC{69 MB} in size. 
While of moderate size compared to large perception DNNs, they are too large
to run on low-power accelerators like the Edge TPU, which 
runs only compressed DNNs that fit in 8MB of on-chip memory.

In our experiments, the \MPNET encoder runs on the robot, but the \textit{pre-trained} planning network runs on a remote server to potentially plan paths for a robotic swarm. Crucially, rather than transmitting a large obstacle embedding $z_0$ as in the original \MPNET, we insert a small encoder at the robot which generates a minimal representation $z \ll z_0$, which is sent across a wireless link, decoded to $z_0$, and passed into the planning network to generate a motion plan. As described in Alg. \ref{alg:train}, we only learn the encoder and decoder to generate compressed environment encoding $z$ and use the original \MPNET planning network. 

Fig. \ref{fig:mars_generalization} shows the percentage of collision-free, feasible paths (y-axis) generated by our \name co-designed planner and the original \MPNET on the same 110 test environments and \SC{40K} test paths provided by the \MPNET authors.
In all cases, if a feasible, collision-free path is found, its path length is within \SC{1\%} of the path generated by the  RRT$^{*}$ planner. Our key result is that \name achieves collision-free paths with costs and failure rate within \SC{1\%} of the original \MPNET, but with much smaller environment encodings $z$. For example, \name achieves good performance for \revision{$z=4$ as opposed to $z=28$ for the original \MPNET, leading to a compression gain of $7 \times$}.

\begin{figure}
    \centering
    \includegraphics[width=1.0\columnwidth]{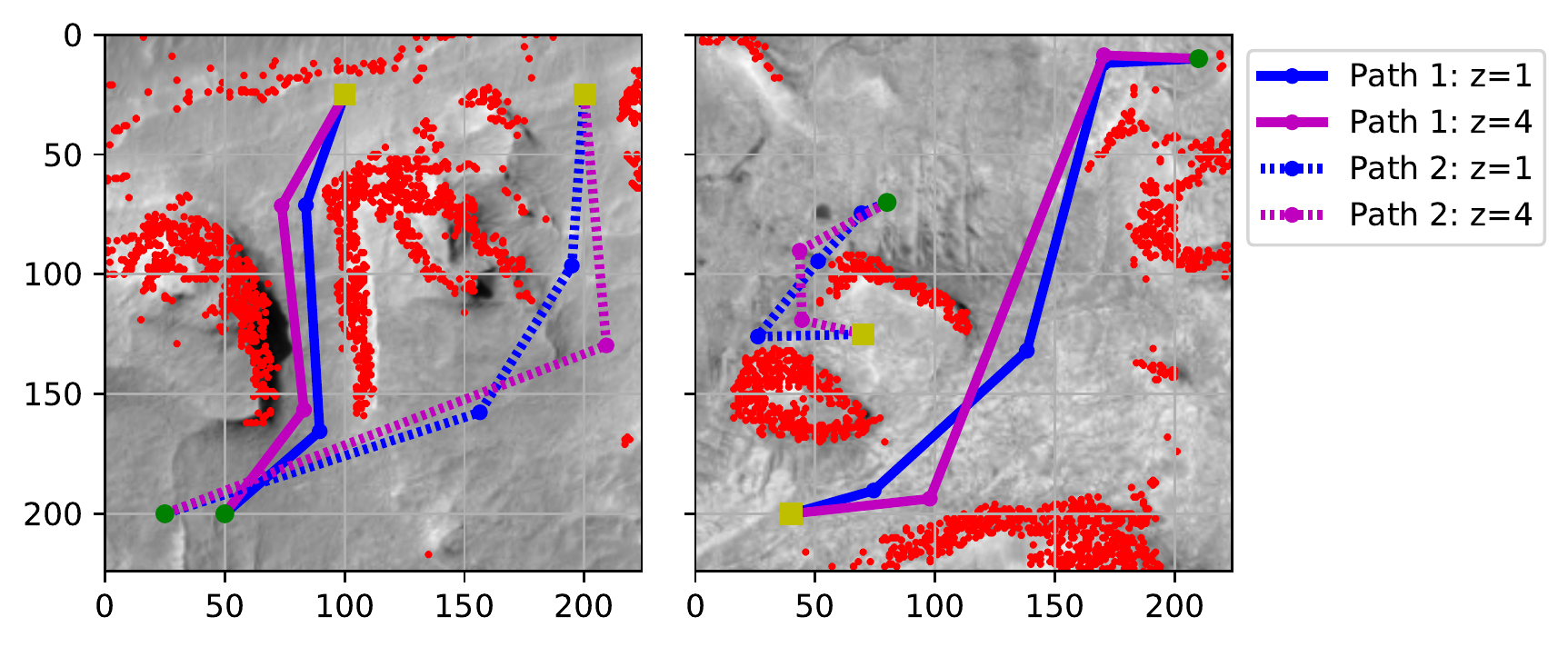}
    \caption{\textbf{Neural Motion Planning on Mars Terrain Data:}
    Our task-aware motion planner generalizes to unseen Martian terrain data for much smaller
    point-cloud embedding sizes $z$ compared to a publicly-available neural motion planner.
    } 
    \label{fig:mars_motion_plan}
    \vspace{-1em}
\end{figure}

\tu{Generalization of motion planner to unseen domains:}
\revision{We further tested the \name motion planner on obstacle point clouds from the Mars HiRise dataset, where the red point clouds in Fig. \ref{fig:mars_motion_plan} represent regions above 30 degrees of elevation}. Even though \MPNET and \name's train and test environments had \revision{point clouds representing \textit{polygonal} obstacles}, \name adapts to curved contours of \textit{never-before-seen} Mars terrain with small bottlenecks $z$.

Fig. \ref{fig:mars_generalization} validates the above observation on over 40.5K paths, where both \MPNET and \name performed similarly on the first two scenarios of seen and unseen environments in \MPNET's original test data. Specifically, a seen test environment was observed by the planner during training, albeit with different start and end test configurations \cite{qureshi2019motion}. Our key experimental result is shown in the third column of Fig. \ref{fig:mars_generalization}, where \name significantly outperforms the original \MPNET for small bottlenecks $z=4,8$ on out-of-domain Mars terrain data. \revision{We hypothesize that \name generalizes better for small $z$ since it focuses the limited information capacity of the bottleneck to represent salient features for planning, and sacrifices reconstruction loss, which is irrelevant to the task goal. Since we only had access to a large, out-of-domain dataset for the Mars planning example, we plan to further stress test \name's generalization capabilities in future work}. 
We plot all benchmarks' performance in our online report \cite{tasknet}, since the trends match Figs. \ref{fig:LQR_simulation_3D}, \ref{fig:MNIST_task_recon_loss}, and \ref{fig:IoT_task_recon_loss}. 

\subsection{Environmental sensor timeseries anomaly detection}
\label{subsec:IoT}
\begin{figure}
    \centering
    \subfloat{
		\centering
		\includegraphics[width=0.499\columnwidth]{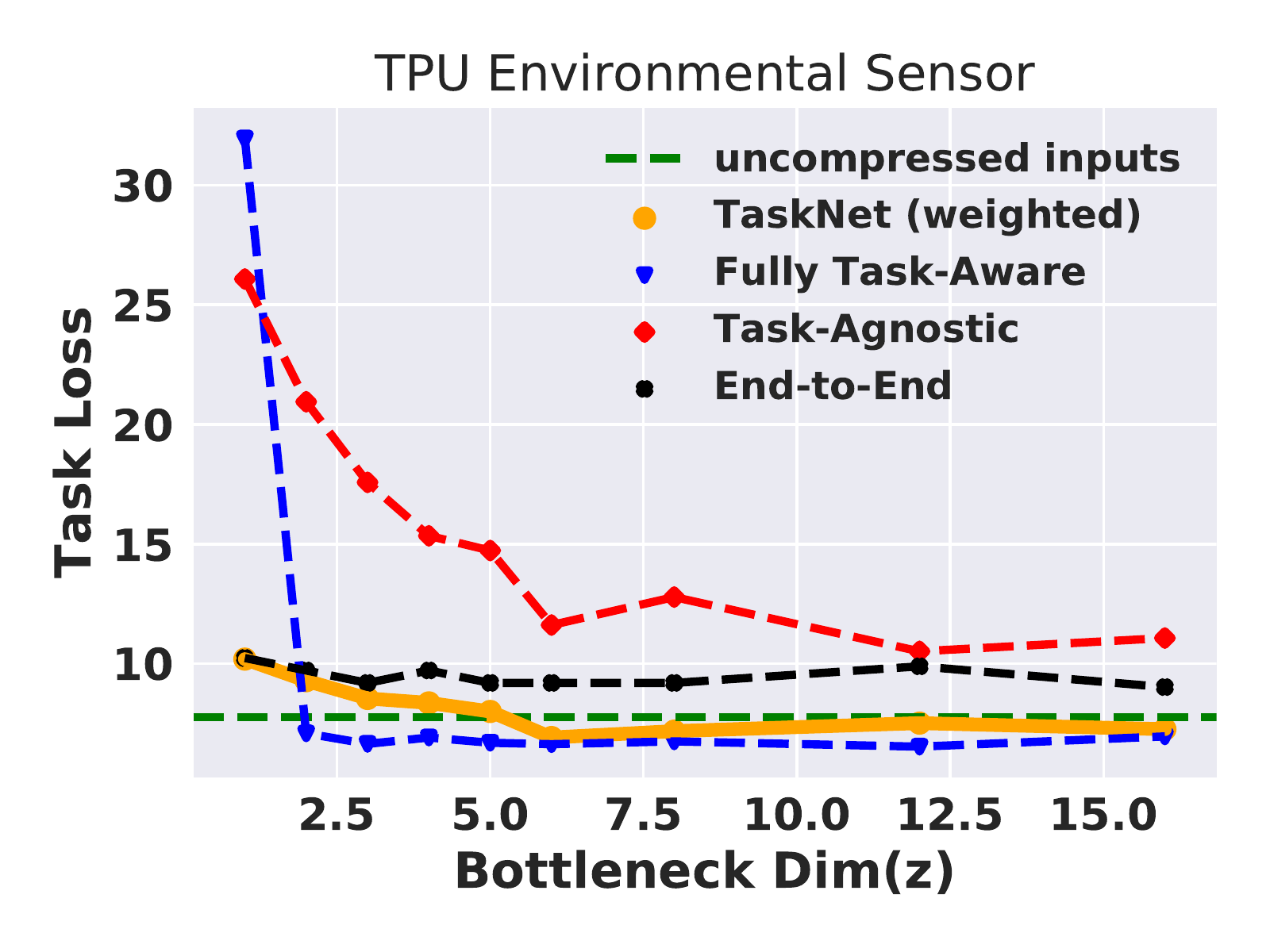}
        \label{fig:IoT_task_loss}
	}
    \subfloat {
        \includegraphics[width=0.499\columnwidth]{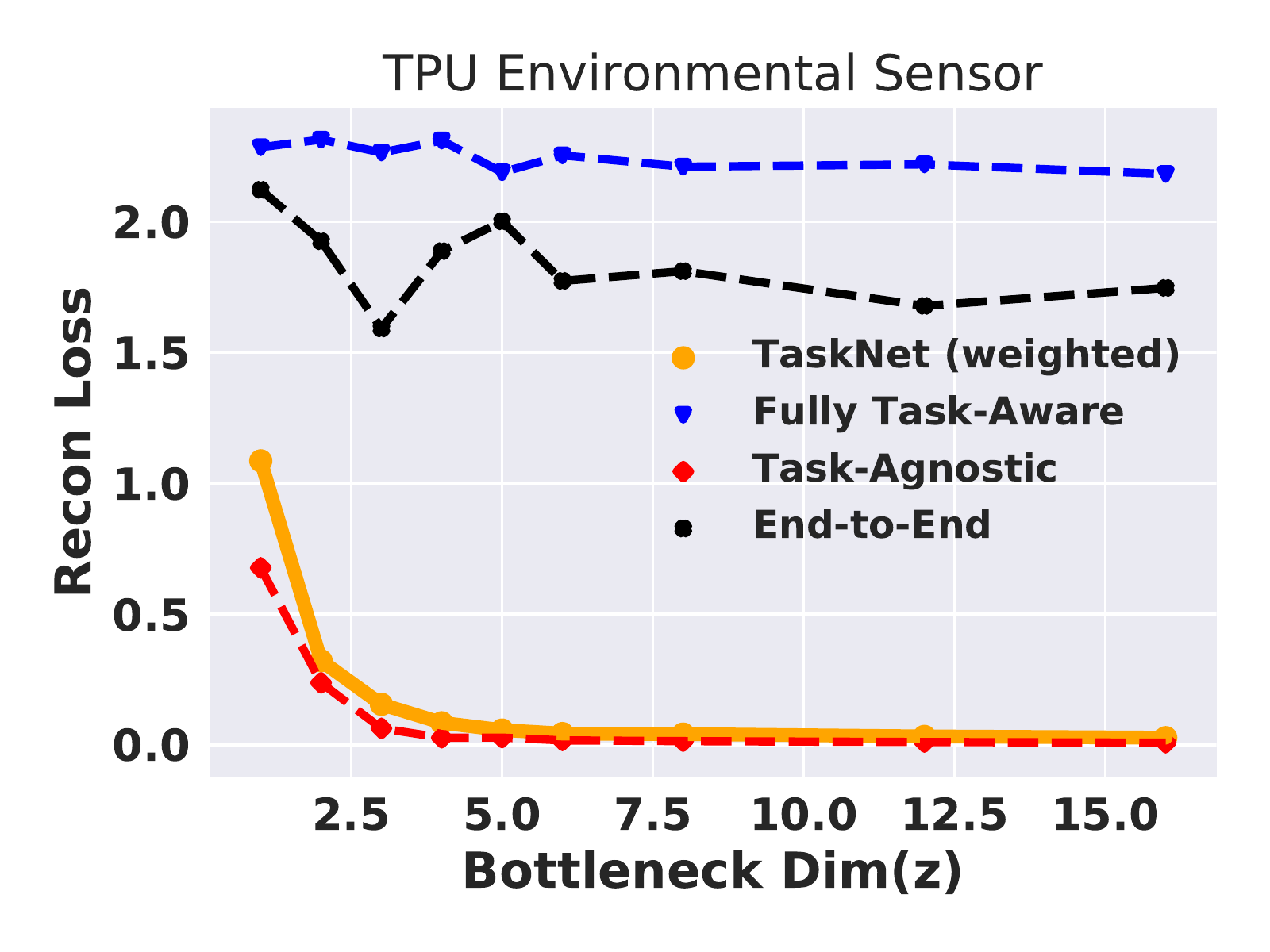}
        \label{fig:IoT_recon_loss}
	}
    \caption{\textbf{Environmental Sensing:} \name achieves low reconstruction loss and matches the low classification error rate achieved when using \textit{original} sensor data.}
    \label{fig:IoT_task_recon_loss}
    \vspace{-1em}
\end{figure}

To highlight how \name generalizes to multiple sensor modalities, we emulated the scenario of a micro-robot, potentially part of a swarm, that streams environmental
sensory data to a remote server for anomaly detection. We used an environmental sensor that connects to the Edge TPU to measure light, temperature, pressure, and humidity timeseries, which are shown in our online report \cite{tasknet}.
The task network $\ftask(;\thetatask)$ is a compact neural network that classifies each timeseries window of length $w$, $x^{t-w:t}$, into three sensor conditions $y^t$. The first class $y^t = 0$ represents if the sensor is being tightly clamped and tampered with, which leads to a rapid fluctuation in the light and humidity measurements. The second class $y^t = 1$ represents when the sensor reads a temperature spike, for which we generate training data by placing a hot hair-dryer over the sensor. The third class $y^t = 2$ represents natural environmental variation without anomalies. 
We collected two weeks of sensor data for a total of 30 training and 30 test traces of 5 minutes each, equally balanced across all three classes.
The task network $\ftask(;\thetatask)$, which was trained on stochastic, diverse timeseries, achieved a promising \SC{90\%} test accuracy. 
Fig. \ref{fig:IoT_task_recon_loss} shows that \name aggressively compresses the timeseries measurements and outperforms the task-agnostic benchmark.
Overall, our diverse results show the promise of provisioning publicly-available task modules at a central server and co-designing minimal representations for networked perception.

\tu{Limitations of our work:}
Our current method requires a constraint on the maximum bottleneck $\zbottleneck$, based on wireless network capacity, and performs a search over increasing sizes $z$. A promising future direction is to automatically determine the optimal representation dimension $\zbottleneck^{*}$ that minimizes weighted loss. Further, we could extend our experiments to compress LIDAR point clouds and long segments of video.

\section{Discussion and Conclusions}
\label{sec:conclusion}
This paper presents a novel framework to aggressively compress rich robotic sensory data for the ultimate needs of a \textit{machine} sensing task, which allows robots to reduce on-board computation and communication bandwidth by up to \SC{11 $\times$}. 
We envision that our co-design algorithm can be applied to a variety of resource-constrained robots that need to distill sensory data for remote inference, such as future Mars Rovers, microrobots, and even low-power drones that collaborate by communicating over 5G wireless networks.

In future work, we plan to extend our algorithm to create minimal, task-relevant representations for cooperative control and develop representations that generalize across several tasks, leveraging ideas from multi-task and meta-learning.
While our approach reduces on-board computation, we plan to also quantify the power consumption of communicating minimal representations. Finally, given our promising results with EfficientNets and deep learning accelerators, we plan to deploy our algorithm on a networked robotic autonomy stack which supports remote inference and tele-operation.

\bibliographystyle{abbrv}
\bibliography{ref/ms}

\newpage
\appendix
\label{sec:appendix}
In this section, we provide further experimental results to illustrate the benefits of co-designing sensory representations with a pre-trained task module's objective using our \name algorithm. Specifically, we elaborate on results for:

\begin{enumerate}
    \item linear task-aware compression (Sec. \ref{appendix:linear})
    \item martian terrain classification with EfficientNets (Sec. \ref{appendix:terrain})
    \item neural motion planning (Sec. \ref{appendix:motion_plan})
    \item anomaly detection of environmental sensor timeseries (Sec. \ref{appendix:environmental_sensor}), and
    \item decoding human-interpretable reconstructed sensory inputs (Sec. \ref{appendix:MNIST}).
\end{enumerate}

\subsection{Further analysis of Linear Task-aware compression}
\label{appendix:linear}
Consider the linear setting described in Sec. \ref{subsec:linear_setup}, where
the encoder is a matrix $\Aencoder \in \reals^{\zbottleneck \times \ninput}$ and the decoder is also a matrix $\Bdecoder \in \reals^{\ninput \times \zbottleneck}$. Recall that the original sensory input is given by $x \in \reals^\ninput$ and reconstructed sensory input is given by $\hat{x} = BAx$, while the task output is denoted by $y = \KLQR x$. \\

We now consider the weighted loss function with $\lambda > 0$, where
\begin{align*}
    \weightedloss = \Vert (\KLQR x - \KLQR \Bdecoder \Aencoder x) \Vert^2_2 + \lambda \Vert(\reconoperand)\Vert^2_2. \\
\end{align*}

In the linear setting, solving Problem \ref{prob:codesign} for a fixed task matrix $K$ amounts to setting the gradient of the weighted task loss with respect to the encoder and decoder matrices to zero. 
Thus, the optimal encoder/decoder matrices that minimize weighted task loss must satisfy:

\begin{align*}
    \nabla_{\Aencoder} \weightedloss &= -2 \Bdecoder^T \KLQR^T (\taskoperand) x^T \\ & -2 \lambda \Bdecoder^T (\reconoperand) x^T \\
    \nabla_{\Bdecoder} \weightedloss &= -2 \KLQR^T (\taskoperand) x^T \Aencoder^T \\ &- 2 \lambda (\reconoperand) x^T \Aencoder^T 
\end{align*}

\begin{figure}
    \centering
    \subfloat{
    \includegraphics[width=0.49\columnwidth]{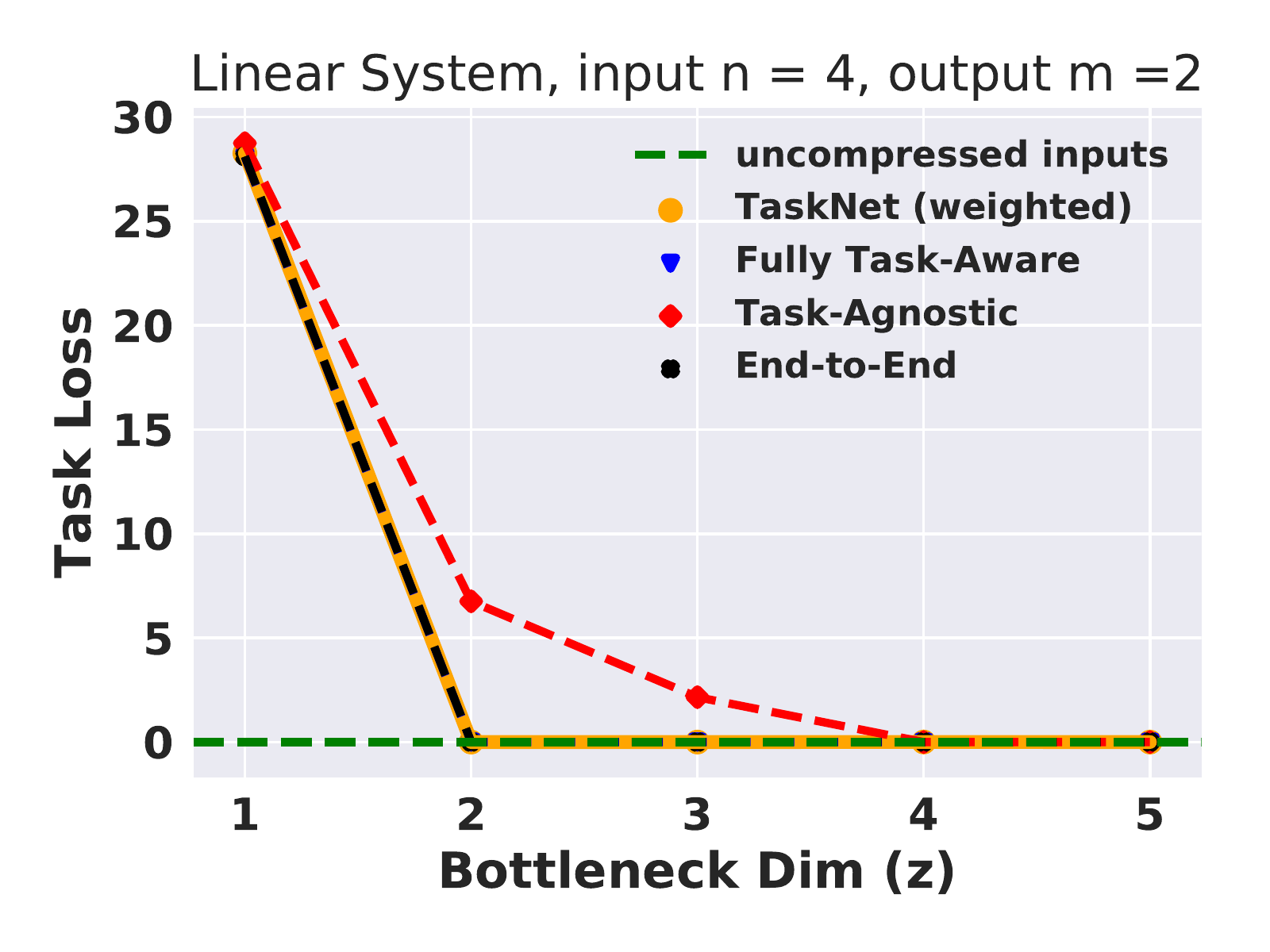} 
    \label{fig:LQR_2D_task_loss}
    }
    \subfloat {
        \includegraphics[width=0.49\columnwidth]{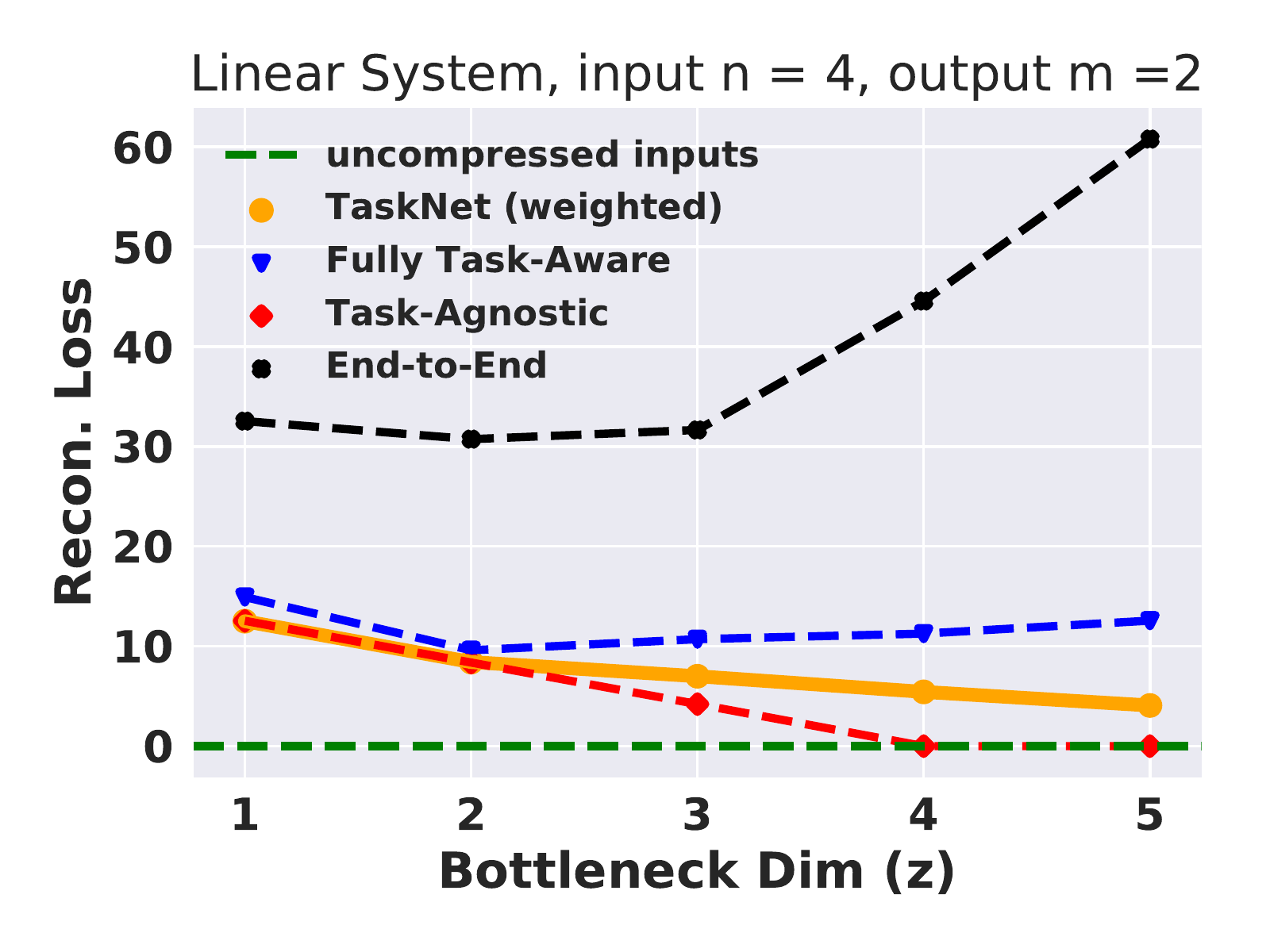} 
    \label{fig:LQR_2D_recon_loss}
    }
    \caption{\textbf{Illustrative example of the benefits of task-aware compression.} (Left, a) Even when the encoder, decoder, and task module are matrices with a bottleneck $z$, optimizing for task loss yields better task accuracy for smaller $z$, which we can analytically prove for the simplified linear setting. (Right, b) However, there is a tradeoff with reconstruction loss, prompting our use of a weighted reconstruction and task loss.}
    \label{fig:LQR_simulation}
\end{figure}

If we run our co-design algorithm Alg. \ref{alg:train} and use the \textit{analytical} gradients computed above in the backpropagation step, we arrive at the results in Figs. \ref{fig:LQR_simulation_3D} and \ref{fig:LQR_simulation}. Specifically, we see the task-aware weighted scheme achieves the upper bound task accuracy with low reconstruction loss for $\zbottleneck \ge \text{rank}(K)$. Further, the relative merits of our weighted task-aware scheme mirror the trends shown in the large-scale deep learning experiments.

\subsection{Further Description of Mars Terrain Classification with EfficientNet}
\label{appendix:terrain}
\begin{figure}
    \centering
    \includegraphics[width=1.0\columnwidth]{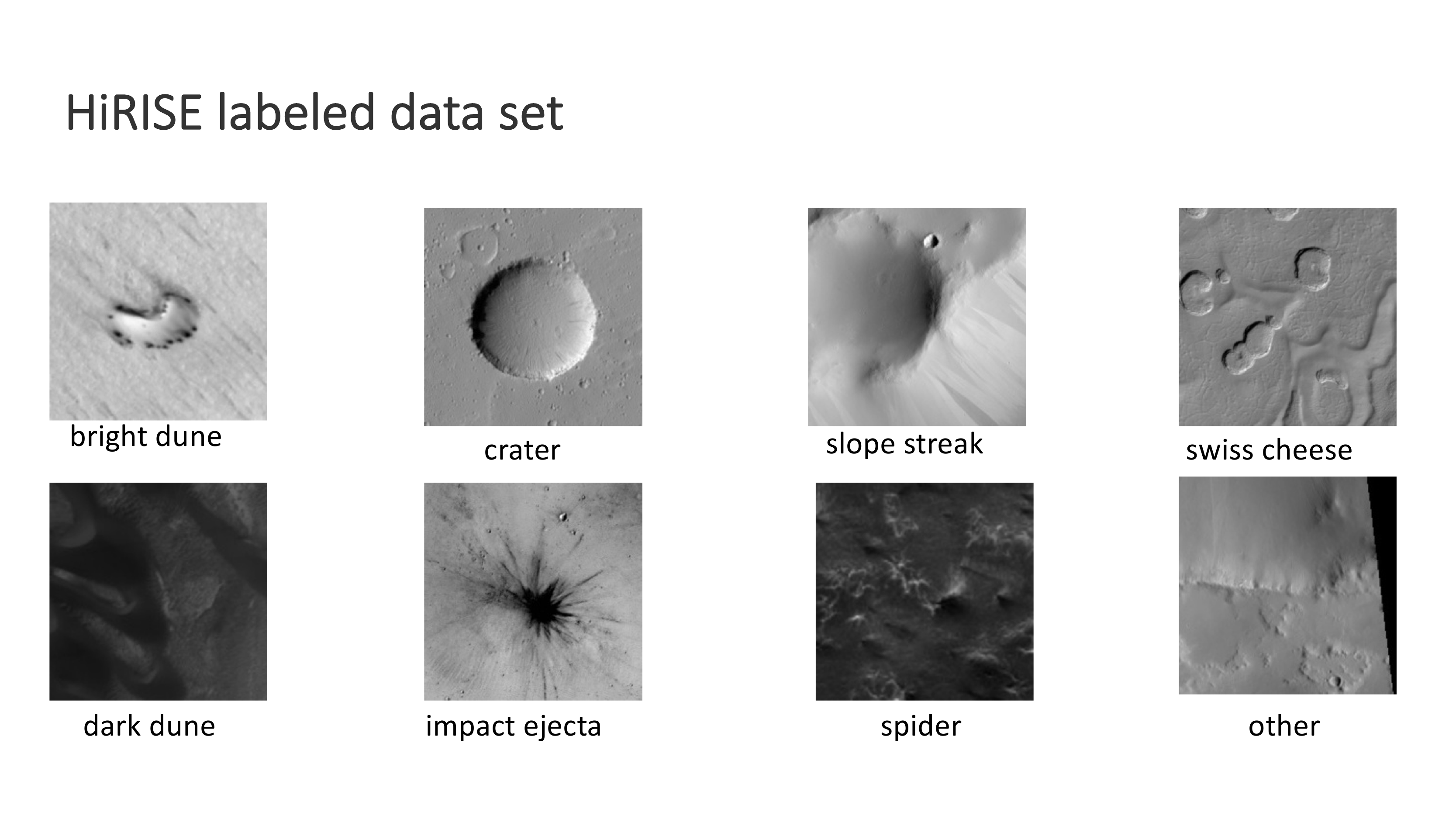} 
    \caption{\textbf{Mars HiRise terrain images \cite{hirise}:} Example labeled Martian terrain images from \cite{hirise}, used for the EfficientNet classification experiments in Sec. \ref{subsec:mars}.} 
    \label{fig:mars_hirise_labels}
\end{figure}

\begin{figure}
    \centering
    \includegraphics[width=1.0\columnwidth]{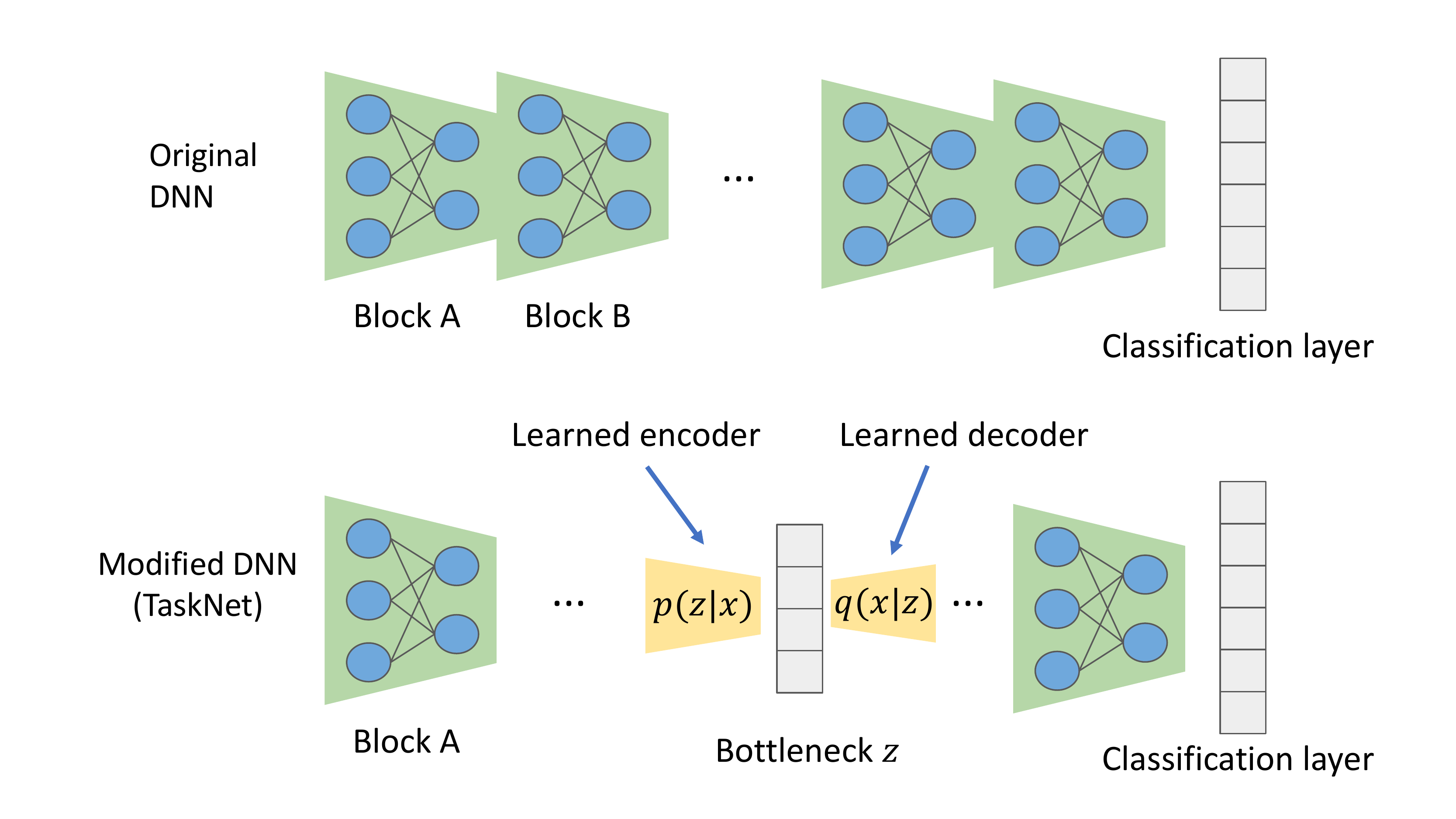} 
    \caption{\textbf{Flexible allocations of robot and server compute:} An example configuration for flexibly allocating compute between a robot and server, used for the EfficientNet Mars classification examples in Sec. \ref{subsec:mars}. Specifically, we can take the \textit{pre-trained} EfficientNet and insert a learned, task-relevant encoder to map an intermediate feature map to a bottleneck representation $z$. Once $z$ is transmitted over a wireless link, the learned decoder uncompresses the intermediate feature map so that computation can continue at a remote server with the rest of the pre-trained EfficientNet. Our flexible co-design algorithm also allows the modular EfficientNet to fully run at a remote server.}
    \label{fig:tasknet_architecture}
\end{figure}

\begin{figure}
    \centering
    \includegraphics[width=1.0\columnwidth]{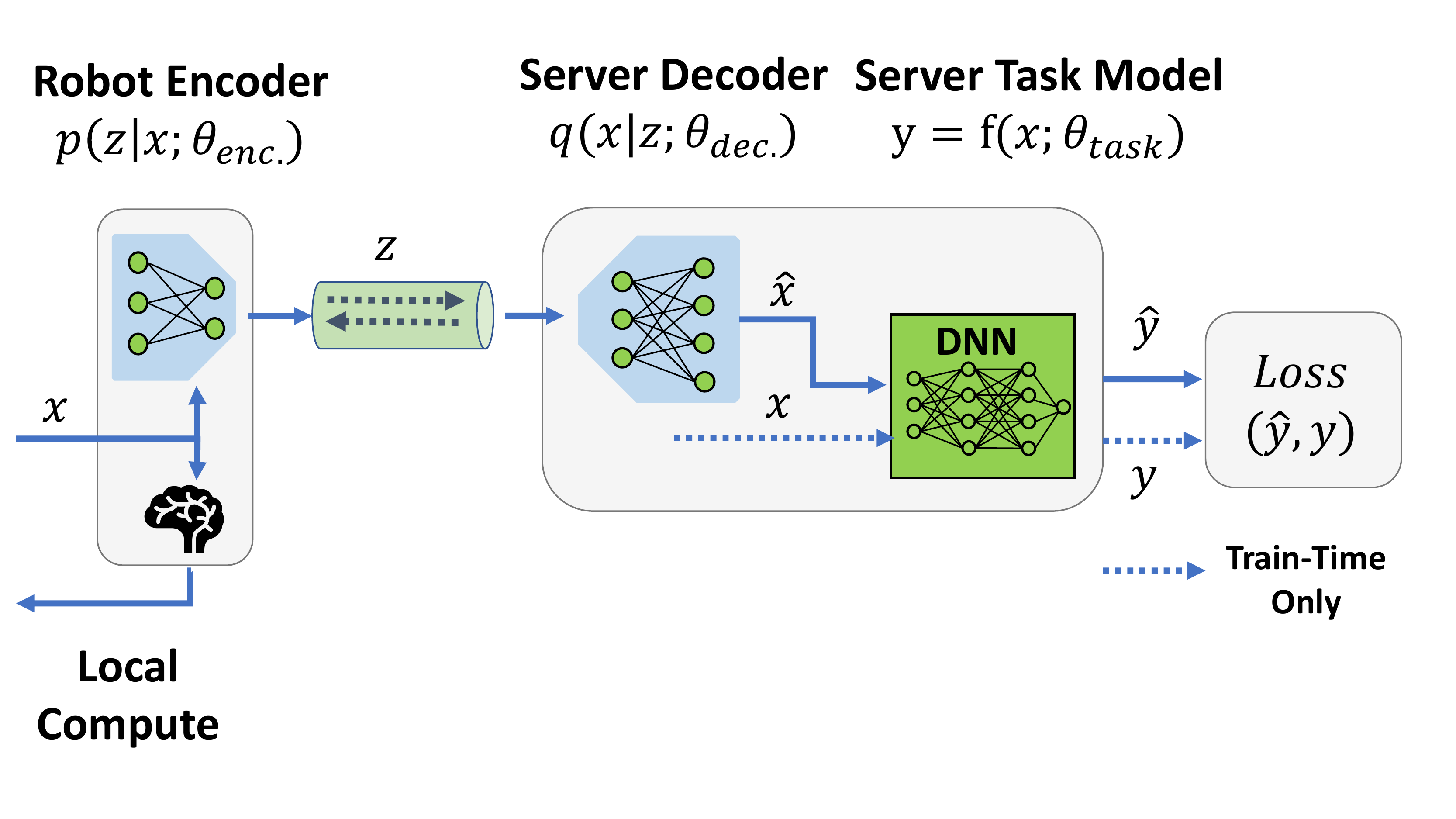}
    \caption{\textbf{Remote perception configuration:} This figure illustrates the conventional configuration of \name, where the robot's learned encoder maps sensory input $x$ to compressed representation $z$. After $z$ is transmitted over a network link, it is decoded to $\xhat$ at a server and directly passed to a pre-trained, modular task network $\ftask(;\thetatask)$ that fully runs at a remote server. This configuration corresponds to Fig. \ref{fig:task_autoencoder}, which complements the alternative architecture in Fig. \ref{fig:tasknet_architecture} where \name flexibly allocates computation that is split between a robot and server.}
    \label{fig:modular_compute}
\end{figure}

\begin{figure}
    \centering
    \includegraphics[width=1.0\columnwidth]{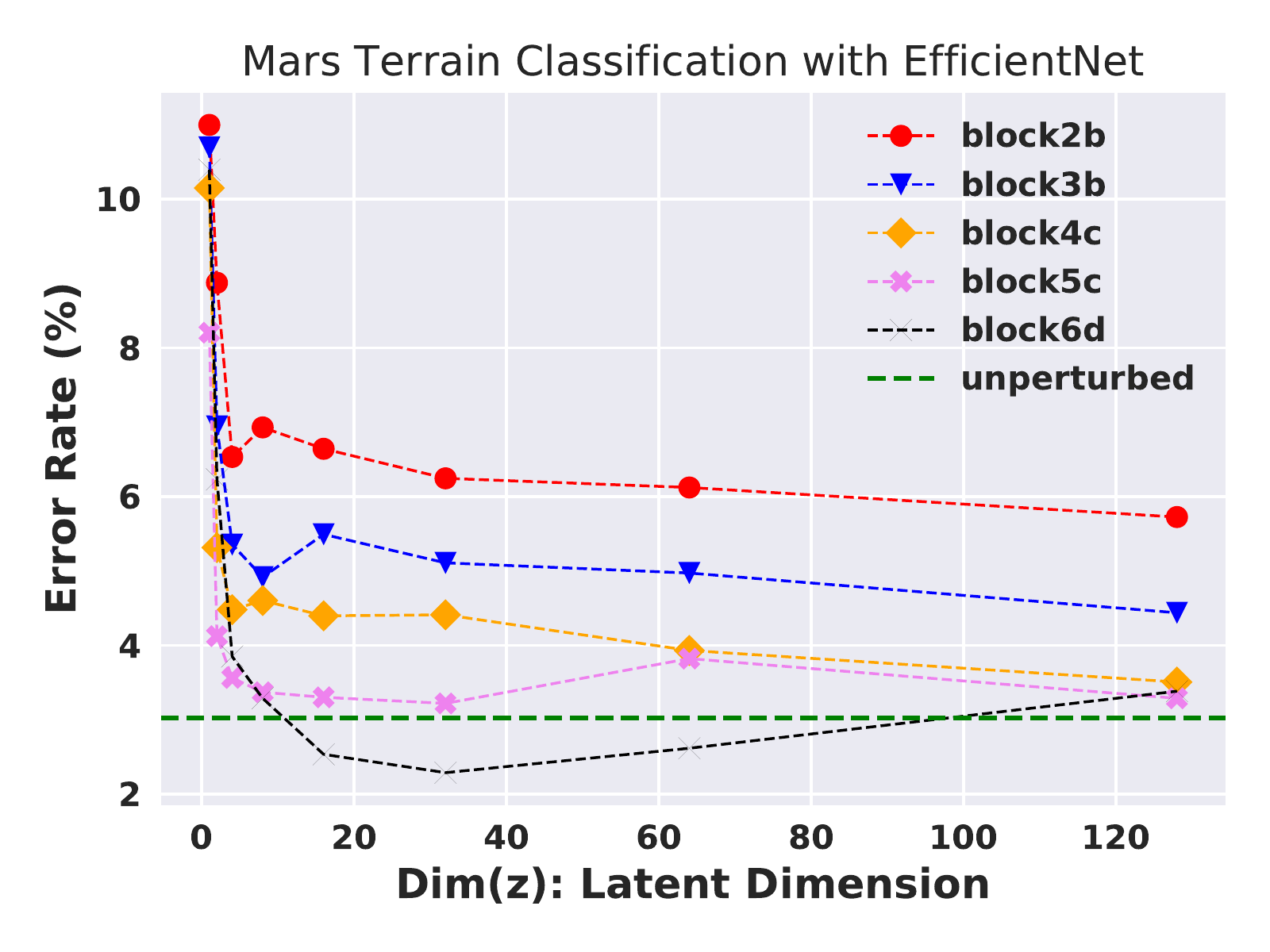} 
    \caption{\textbf{Mars terrain classification:} This plot illustrates our ability to flexibly allocate compute between a resource-constrained robot and cloud server. Specifically, as described in Sec. \ref{subsec:mars}, we can run layers until EfficientNet block $b$, co-design task relevant encoder/decoder to transmit intermediate computations, and finally continue computation with the rest of the \textit{pre-trained} EfficientNet at a server. We plot \name with $\lambda=0$ for scenarios where the robot only runs up to EfficientNet block $b$ locally and achieve within $1\%$ of the uncompressed input error rate (green dashed line) at block $4c$ and beyond for $\zbottleneck=128$.}
    \label{fig:mars_hirise_all_splits}
\end{figure}

In Sec. \ref{subsec:mars}, we showed how to flexibly allocate compute resources between a resource-constrained robot and server for terrain classification with EfficientNets. Fig. \ref{fig:mars_hirise_labels} illustrates the training data from the HiRise dataset \cite{hirise}. Then, Fig. \ref{fig:tasknet_architecture} illustrates how we can flexibly allocate compute between a small robot and server with the \name architecture. However, we emphasize that our design also allows the full, modular EfficientNet task network to be run remotely, as depicted in the complementary configuration in Fig. \ref{fig:modular_compute}. Finally, Fig. \ref{fig:mars_hirise_all_splits} illustrates the resultant classification task accuracy for various configurations of computation allocated between a robot and server.

\subsection{Generalization of Neural Motion Planning to out-of-domain Mars Data}
\label{appendix:motion_plan}
\begin{figure}
    \centering
    \includegraphics[width=1.0\columnwidth]{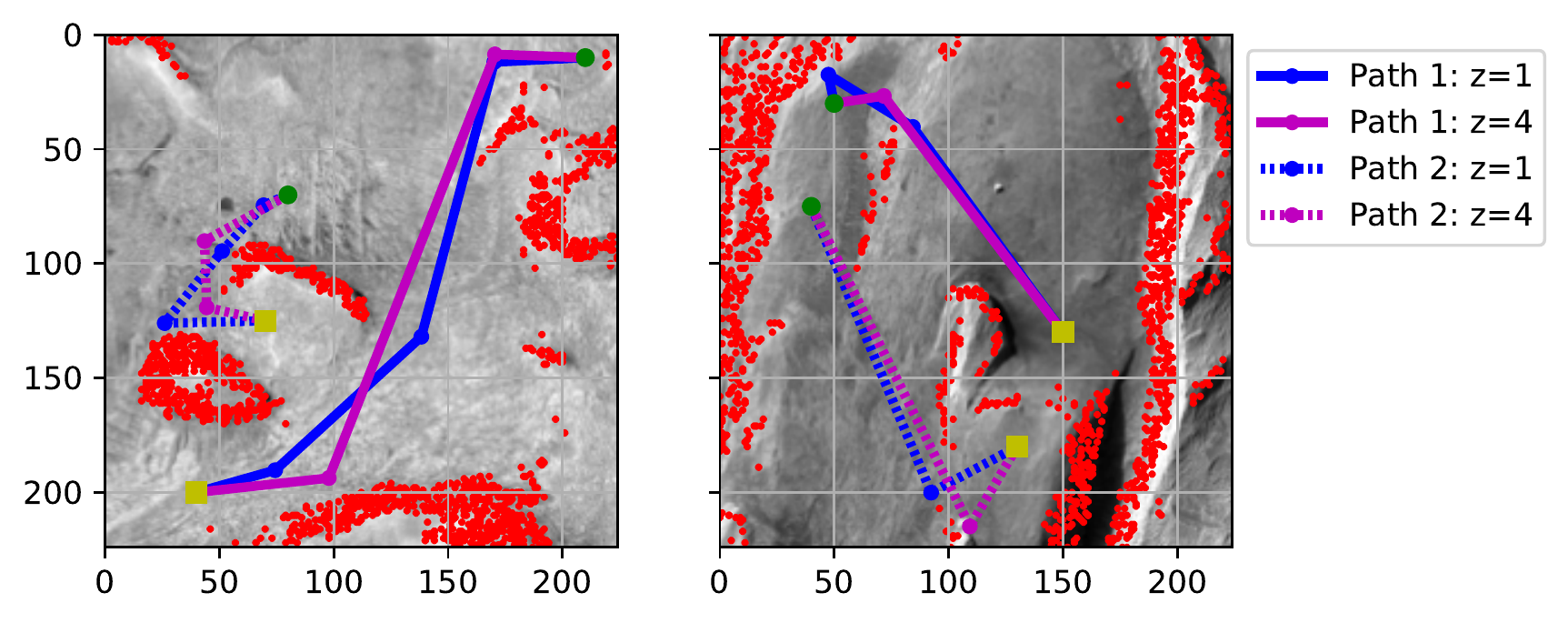} 
    \caption{\textbf{Motion planning on Mars Terrain Data}: A further example of how \name can plan feasible, collision-free motion plans on out-of-domain Mars data, as illustrated in Fig. \ref{fig:mars_generalization} and \ref{fig:mars_motion_plan}.}
    \label{fig:mars_motion_plan_2}
\end{figure}

In Section \ref{subsec:motion_planning} and Figs. \ref{fig:mars_generalization} and \ref{fig:mars_motion_plan}, we showed that \name generalizes to never-before-seen Mars terrain data, where obstacle point clouds have significantly different shapes than those of the \MPNET training and test data. Further examples of the \name co-designed motion planner are provided in Fig. \ref{fig:mars_motion_plan_2}.

\subsection{Further Description of Environmental Sensor Experiments}
\label{appendix:environmental_sensor}

\begin{figure}
    \centering
    \includegraphics[width=0.99\columnwidth]{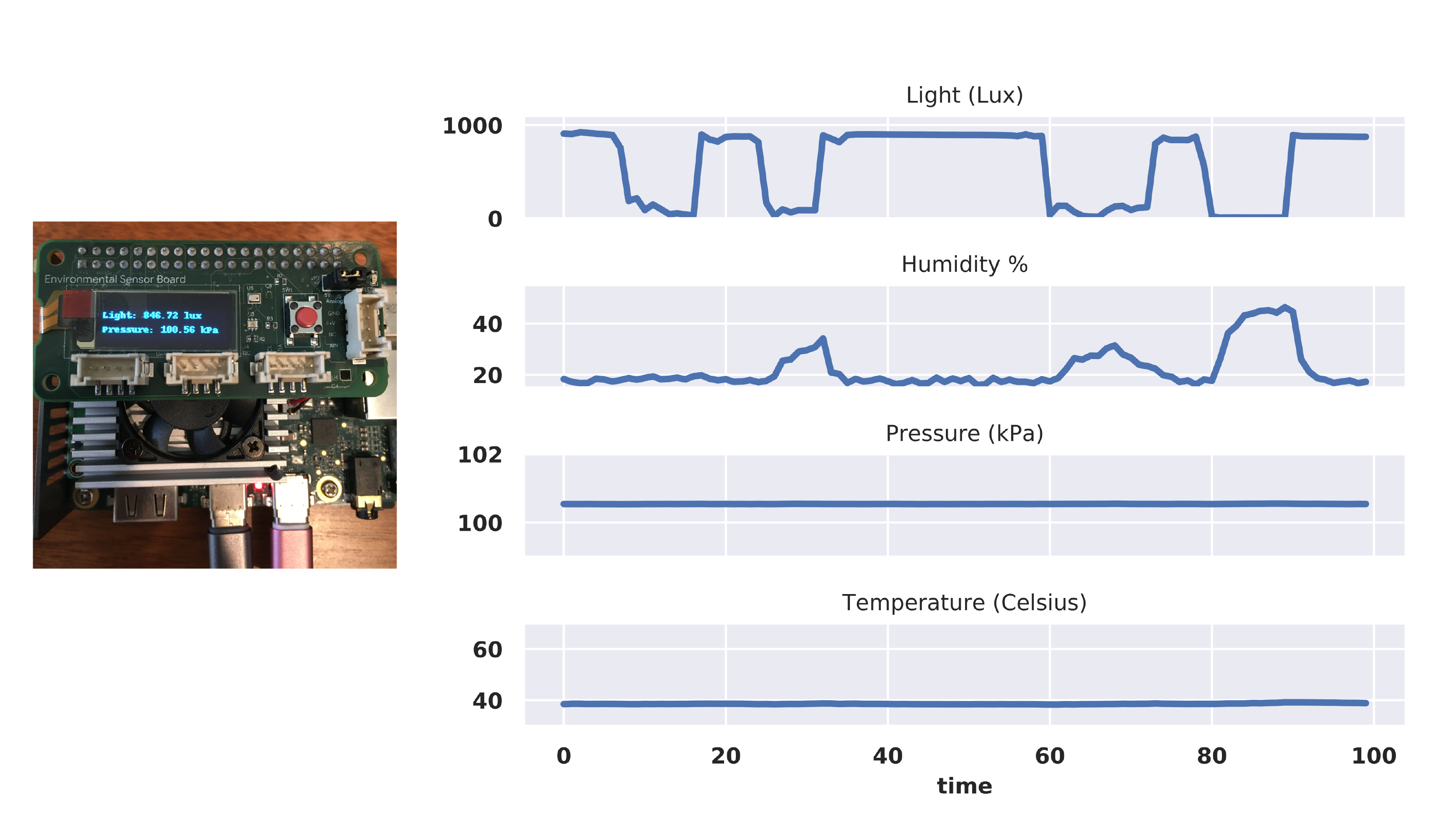} 
    \caption{(Left) Environmental sensor on the Google Edge Tensor Processing Unit (TPU) and example stochastic timeseries measurements (right). Our experiments in Sec. \ref{sec:experiments} illustrate how \name can compress these sensory measurements for anomaly detection using a neural network task module.}
    \label{fig:IoT}
\end{figure}

Fig. \ref{fig:IoT} shows example timeseries that we collected from an environmental sensor board connected to the Edge TPU DNN accelerator. Our comprehensive dataset spans two weeks of diverse, stochastic timeseries described in Sec. \ref{subsec:IoT}.

\subsection{Effects of reconstruction loss on decoded images}
\label{appendix:MNIST}
\begin{figure}
    \centering
    \includegraphics[width=1.0\columnwidth]{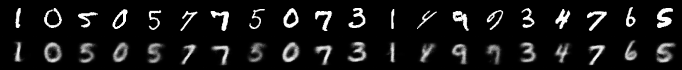}
    \caption{\textbf{MNIST with reconstruction loss}: Our general co-design framework of \name allows highly-compressible, but also human-interpretable decoded images. Specifically, when we optimize for classification task loss and weighted reconstruction loss, with weight $\lambda=0.01$ (Eq. \ref{eq:weighted_loss}), decoded images $\xhat$ (bottom row) are human-interpretable and resemble original images $x$ (top row). However, the key benefit of \name is that we have a heavily-compressed bottleneck representation of $\zbottleneck=4$, which is much smaller than task-agnostic representations, as shown in Fig. \ref{fig:MNIST_task_recon_loss}. We note that the decoded samples emphasize salient features for machine classification and do not produce the sharpest reconstructions since reconstruction is not the \textit{primary} objective. However, these blurry decoded samples $\xhat$ are sufficient to achieve high \textit{machine} classification accuracy.}
    \label{fig:mars_recon_weighted}
\end{figure}

\begin{figure}
    \centering
    \includegraphics[width=1.0\columnwidth]{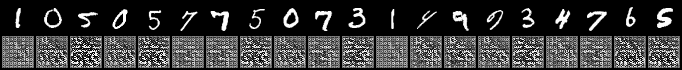}
    \caption{\textbf{MNIST classification \textit{without} reconstruction loss}: 
    Our general co-design framework also allows us to create highly-compressible representations that are purely optimized for a machine's task loss with reconstruction weight $\lambda=0.0$ (Eq. \ref{eq:weighted_loss}). Though we have a heavily-compressed representation with a size of $\zbottleneck=4$, decoded images $\xhat$ (bottom row) are \textit{not} human-interpretable and don't resemble original images $x$ (top row), unlike the scenario of Fig. \ref{fig:mars_recon_weighted}. This motivates our addition of a regularization term with reconstruction weight $\lambda > 0$ to develop representations that are highly-compressible, but yield interpretable decoded samples $\xhat$ as in Fig. \ref{fig:mars_recon_weighted}.}
    \label{fig:mars_recon_unweighted}
\end{figure}

\begin{figure}
    \centering
    \includegraphics[width=1.0\columnwidth]{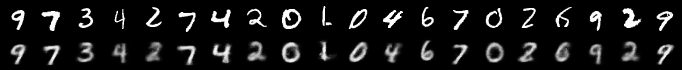}
    \caption{\textbf{MNIST with reconstruction loss}: 
    A further example of how our general co-design framework of \name allows highly-compressible, but also human-interpretable decoded images. This example mirrors Fig. \ref{fig:mars_recon_weighted}, but for another set of data.}
    \label{fig:mars_recon_weighted_2}
\end{figure}

\begin{figure}
    \centering
    \includegraphics[width=1.0\columnwidth]{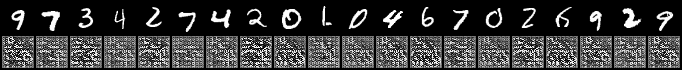}
    \caption{\textbf{MNIST \textit{without} reconstruction loss}: 
    A further example of how optimizing for purely a machine's task loss, without reconstruction weight $\lambda$, yields highly-compressible, but uninterpretable decoded images $\xhat$. This example mirrors Fig. \ref{fig:mars_recon_unweighted}, but for another set of data.}
    \label{fig:mars_recon_unweighted_2}
\end{figure}

In Eq. \ref{eq:weighted_loss}, we allow a roboticist to optimize for task loss, but also optionally for reconstruction loss, in scenarios where decoded inputs $\xhat$ need to be sanity-checked or viewed remotely to debug. For the MNIST experiments in Sec. \ref{subsec:MNIST}, we show how introducing a reconstruction loss weight $\lambda > 0$ yields highly-compressible, but human-interpretable results in Figs. \ref{fig:mars_recon_weighted} and \ref{fig:mars_recon_weighted_2}. We contrast these with the case where we optimize \textit{solely} for a machine's task loss in Figs. \ref{fig:mars_recon_unweighted} and \ref{fig:mars_recon_unweighted_2} with $\lambda=0$.

\label{appendix:MNIST_recon}

\end{document}